\documentclass[noblindrev,msom]{informs3noheader} 
\usepackage{booktabs}
\usepackage{mdframed}
\usepackage{algpseudocode}
\usepackage{algorithm}
\usepackage{color}
\usepackage{bbm}
\usepackage{tabularx}
\usepackage{rotating}
\usepackage{graphicx}
\usepackage{subcaption}
\usepackage{caption}
\RequirePackage{endnotes}
\usepackage{soul}
\usepackage{listings}
\usepackage{xcolor}

\usepackage{refcount}

\definecolor{codegreen}{rgb}{0,0.6,0}
\definecolor{codegray}{rgb}{0.5,0.5,0.5}
\definecolor{codepurple}{rgb}{0.58,0,0.82}
\definecolor{backcolour}{rgb}{0.99,0.99,0.95}

\definecolor{mygreen}{rgb}{0,0.6,0}
\definecolor{mygray}{rgb}{0.5,0.5,0.5}
\definecolor{mymauve}{rgb}{0.58,0,0.82}

\lstdefinelanguage{cfg}
{
    basicstyle=\ttfamily\small,
    columns=fullflexible,
    morecomment=[s][\color{mymauve}\bfseries]{[}{]},
    morecomment=[l]{\#},
    morecomment=[l]{;},
    commentstyle=\color{gray}\ttfamily\itshape,
    morekeywords={},
    otherkeywords={=,:},
    keywordstyle={\color{mygreen}\bfseries}
}

\newcolumntype{P}{ >{\arraybackslash} p{.48\columnwidth} }
\newcolumntype{Q}{ >{\arraybackslash} p{.01\columnwidth} }
\newcolumntype{M}{ >{\centering\arraybackslash} m{\columnwidth} }
\newcolumntype{A}{ >{\arraybackslash} p{.44\columnwidth} }
\newcolumntype{B}{ >{\arraybackslash} p{.52\columnwidth} }

\OneAndAHalfSpacedXI

\usepackage{natbib}
\bibpunct[, ]{(}{)}{,}{a}{}{,}%
\def\bibfont{\small}%
\newcounter{note}[section]

\TheoremsNumberedThrough     
\ECRepeatTheorems

\EquationsNumberedThrough    

\MANUSCRIPTNO{}

\usepackage{multirow}
\usepackage{booktabs}
\usepackage{graphicx}
\usepackage{algorithm}
\usepackage{algpseudocode}
\graphicspath{{figures/}}

\begin{document}


	\RUNAUTHOR{Harsha et al.}

	\RUNTITLE{Deep Policy Iteration for Inventory Management}

	\TITLE{Deep Policy Iteration with Integer Programming for Inventory Management
	} 

	\ARTICLEAUTHORS{%
 		\AUTHOR{Pavithra Harsha, Ashish Jagmohan, Jayant Kalangnanam, Brian Quanz}
 		\AFF{IBM Research, Thomas J. Watson Research Center, Yorktown Heights, NY 10598, USA}
 		\AUTHOR{Divya Singhvi}
 		\AFF{NYU Leonard N. Stern School of Business, NY 10012, USA}
}

\ABSTRACT{\textbf{Problem Definition:} In this paper, we present a Reinforcement Learning (RL) based framework for optimizing long-term discounted reward problems with large combinatorial action space and state dependent constraints. These characteristics are common to many operations management problems, e.g., network inventory replenishment, where managers have to deal with uncertain demand, lost sales, and capacity constraints that results in more complex feasible action spaces. Our proposed Programmable Actor Reinforcement Learning (PARL) uses a deep-policy iteration method that leverages neural networks (NNs) to approximate the value function and combines it with mathematical programming (MP) and sample average approximation (SAA) to solve the per-step-action optimally while accounting for combinatorial action spaces and state-dependent constraint sets. \textbf{Results:} We then show how the proposed methodology can be applied to complex inventory replenishment problems where analytical solutions are intractable. We also benchmark the proposed algorithm against state-of-the-art RL algorithms and commonly used replenishment heuristics and find that the proposed algorithm considerably outperforms existing methods by as much as 14.7\% on average in various complex supply chain settings. \textbf{Managerial Insights:} We find that this improvement in performance of PARL over benchmark algorithms can be directly attributed to better inventory cost management, especially in inventory constrained settings. Furthermore, in the simpler setting where optimal replenishment policy is tractable or known near optimal heuristics exist, we find that the RL based policies can learn near optimal policies. Finally, to make RL algorithms more accessible for inventory management researchers, we also discuss the development of a modular Python library that can be used to test the performance of RL algorithms with various supply chain structures. This library can spur future research in developing practical and near-optimal algorithms for inventory management problems.
}%



	\maketitle

	%
	\section{Introduction}
Over the past decade, inventory and supply chain management have undergone significant transformations. The rise of online e-commerce has introduced complexities and globalization to supply chains, resulting in increasingly interconnected physical flows \citep{young_2022}. Consequently, the costs associated with managing these supply chains have escalated. Additionally, the COVID-19 pandemic has further driven up expenditures in managing these complex networks. According to a recent Wall Street Journal report, U.S. business logistics costs surged by as much as 22\% year-over-year \citep{young_2022}. A promising avenue for addressing these challenges lies in the implementation of Artificial Intelligence (AI)-based inventory management solutions, which offer potential benefits for managing intricate supply chains.

AI and Reinforcement Learning (RL) have achieved remarkable breakthroughs in various domains, including games \citep{mnih2013playing}, robotics \citep{kober2013reinforcement}, and healthcare \citep{yu2019reinforcement}. RL provides a systematic framework for solving sequential decision-making problems with minimal domain knowledge. By leveraging state-of-the-art, open-source RL methods, one can develop effective policies that maximize long-term rewards for many sequential decision problems. It is, therefore, not surprising that RL has been applied to domains such as supply chains \citep{gijsbrechts2018can, oroojlooyjadid2021deep, sultana2020reinforcement}. However, the application of RL to enterprise-level operations remains limited and poses significant challenges.

Typical operations management (OM) problems, such as inventory management and network revenue management, are characterized by large action spaces, well-defined state-dependent action constraints, and underlying stochastic transition dynamics. For instance, a firm managing inventory across a network of nodes in a supply chain must decide how much inventory to allocate to different nodes \citep{caro2010inventory}. This involves overcoming various challenges: dealing with uncertain demand across the network nodes, managing a large set of locally feasible (often combinatorial) actions, ensuring numerous state-dependent constraints are met to maintain feasibility, and balancing the trade-off between immediate and long-term rewards. RL methods rely on both live and simulated environments to sample uncertainties and generate reward trajectories. By evaluating these trajectories, RL can estimate optimal action policies that balance long-term and short-term rewards. However, the large combinatorial action spaces with state-dependent constraints inherent in OM problems make traditional enumeration-based RL techniques computationally infeasible.

In this paper, we present a specialized RL algorithm to address these challenges. Specifically, we introduce a deep-policy iteration method that leverages neural networks (NNs) to approximate the value function. This method combines mathematical programming (MP) and sample average approximation (SAA) to solve per-step actions optimally, accounting for combinatorial action spaces and state-dependent constraints. We employ this modified RL approach to provide benchmark solutions for inventory management problems with complexities that make analytical solutions infeasible, such as lost sales, fixed costs, dual sourcing, and lead times in multi-echelon networks. We then compare our approach with state-of-the-art RL methods and popular supply chain heuristics.

	\subsection{Contributions}
We make the following contributions through this work:
\begin{itemize}
    \item We present a policy iteration algorithm for dynamic programming problems with large action spaces and underlying stochastic dynamics, which we call Programmable Actor Reinforcement Learning (PARL). In our framework, the value-to-go is represented as the sum of immediate reward and future discounted rewards. The future discounted rewards are approximated with a neural network (NN) that is fitted by generating value-to-go rewards from Monte Carlo simulations. We then represent the NN and the immediate reward as an integer program, which is used to optimize the per-step action. As the approximation improves, so does the per-step action, ensuring that the learned policy eventually converges to the unknown optimal policy. This approach overcomes the challenges of enumeration and allows for easy implementation of known contextual state-dependent constraints.
    \item We apply the proposed methodology to the problem of optimal replenishment decisions for a retailer with a representative network of warehouses and retail stores. Focusing on settings where analytical tractability is not guaranteed, we analyze and compare the performance of PARL under different supply chain network structures (multi-echelon distribution networks with and without dual-sourcing, and with lost sales) and network sizes (from a single supplier and three retailers, to up to 20 heterogeneous retailers with multiple intermediary warehouses). We find that PARL is competitive, and in some cases outperforms, state-of-the-art methods, showing a 14.7\% improvement on average across different complex settings studied in this paper. Our numerical experiments provide comprehensive benchmark results in these settings using different RL algorithms (SAC, TD3, PPO, and A2C), as well as commonly used heuristics in various supply chain settings (specifically, base stock policies on an edge and on a serial path). We also perform additional numerical experiments to analyze the structure of the learned RL-based replenishment policies and find that (i) in a simpler back-order setting, the RL policy is near optimal as it also learns an order-up-to policy and (ii) in more complex settings, higher profitability is due to improved cost management across the supply chain network. Finally, in settings where near-optimal inventory heuristics are known (lost sales setting of \citealt{zipkin2008old}, \citealt{xin2021understanding}, \citealt{gijsbrechts2018can}), similar to previous studies, we find that the DRL methods are able to learn near-optimal heuristics.

    \item Finally, we open source our supply chain environment as a Python library to allow researchers to easily implement and benchmark PARL and other state-of-the-art RL methods in various supply chain settings. Our proposed library is modular and allows researchers to design supply chain networks with varying complexity (multi-echelon, dual sourcing), size (number of retailers and warehouses), and different reward structures. While our objective is similar to \cite{hubbs2020or}, we focus on inventory management problems specifically and provide more flexibility in the network design. We believe this library will make RL algorithms more accessible to the OR and inventory management community and spur future research in developing practical and near-optimal algorithms for inventory management problems.
\end{itemize}

	\subsection{Selected Literature Review}\label{subsec:literature_review}

  Our work is related to the following different streams of literature: (1) parametric policies, (2) approximate dynamic programming (ADP), (3) reinforcement learning (RL) and (4) mathematical programming based RL. As the literature in these topics, in general and even within the context of supply chain and inventory management specifically, is quite vast, we describe the state-of-the art with select related literature that is most related to the current work.

\paragraph{Parametric policies for inventory management:} The topic of inventory management has been extensively studied over the years and has a long history. The seminal work of \citet{scarf1960optimality} shows that for a single node sourcing from a single supplier with infinite inventory, a constant lead time, and an ordering cost with fixed and variable components, the optimal policy for back-ordered demand has an $(s,S)$ structure. Here, $S$ is referred to as the order-up-to level based on the inventory position (the sum of on-hand inventory plus that in the pipeline, i.e., a collapsed inventory state), and $s$ is an inventory position threshold below which orders are placed. This $(s,S)$ policy is commonly referred to as the base stock policy. In lost-sales demand settings, where demand in excess of inventory is lost, the structure of the optimal policy is unknown when lead times are non-zero \citep{zipkin2008old, zipkin2008structure}, even in the single node setting sourcing from a single retailer. Moreover, base stock policies generally perform poorly in these settings \citep{zipkin2008old}, except when penalty costs are high, inventory levels are high, and stock-outs are rare. In general, the optimal policy depends on the full inventory pipeline, unlike the state-space collapse possible in back order settings \citep{levi20082}, and the complexity grows exponentially with lead time. In these settings, \citet{huh2009asymptotic} and \citet{goldberg2016asymptotic} respectively prove asymptotic optimality results for the base stock policy when penalties are high and for constant ordering policies when lead times are large. \citet{xin2021understanding} combines these results to propose a capped base stock policy that is asymptotically optimal as lead time increases and shows good empirical performance for smaller lead times. \citet{sheopuri2010new} prove that the lost sales problem is a special case of the dual sourcing problem (where a retail node has access to two external suppliers), and hence base stock policies are generally not optimal. In the dual sourcing setting, various heuristic policies extend the base stock policy with a constant order and/or cap that splits the order between the two suppliers depending on the inventory position across each. These include the Dual-Index \citep{veeraraghavan2008now}, Tailored Base-Surge \citep{allon2010global}, and Capped Dual Index policies \citep{sun2019robust}.

Multi-echelon networks, which have multiple nodes, stages, or echelons holding inventory, present additional challenges. Base stock policies are optimal only in special cases with back-ordered demands without fixed costs, with restrictive assumptions such as a serial chain with a back order penalty at the demand node \citep{clark1960optimal} or the inability to hold demand in the warehouse in a two-echelon distribution network \citep{federgruen1984approximations}. Recently, \citet{bansal2022monge} analyzed a setting where different nodes in the supply chain compete with one another with stockout-based demand substitution. For an extensive review of multi-echelon models based on various modeling assumptions and supply chain network structures, refer to \citet{de2018typology}. Despite the sub-optimality of base stock policies (using $(s,S)$ policies with a collapsed state space via inventory positions), they remain popular both in practice and in the literature. For example, \citet{ozer2008stock} and \citet{rong2017heuristics} propose competitive heuristics that compute the order-up-to base stock levels for multi-echelon distribution (tree) networks without fixed costs but with service level constraints and demand back ordering costs respectively, and show asymptotic optimality in certain dimensions in the two-echelon case. \citet{agrawal2019learning} propose a learning-based method to find the best base stock policy in a single node lost sales setting with regret guarantees. \citet{pirhooshyaran2020simultaneous} develop a DNN-based learning approach to find the best order-up-to levels in each link of a general supply chain network.

In this work, we present a deep RL approach and leverage it to solve a certain class of cost-based stochastic inventory management problems in settings where parametric optimal policies do not exist or are unknown, such as multi-echelon supply chains with fixed costs, capacities, lost sales, and dual sourcing. In these settings, we provide new empirical benchmarks showing that our proposed algorithm outperforms common heuristics based on the base-stock policy.

 \paragraph{Approximate Dynamic Programming (ADP) and Reinforcement Learning (RL):} Our work is also related to the broad field of ADP \citep{bertsekas2012dynamic, powell2007approximate}, which focuses on solving Bellman's equation and has been an area of active research for many years. ADP methods typically use an approximation of the value function to optimize over computationally intractable dynamic programming problems. Popular algorithms can be classified into two types: model-aware and model-agnostic. Model-aware approaches use information on underlying system dynamics (known transition function and immediate reward) to estimate value-to-go, with common approaches including value iteration and policy iteration. These algorithms start with arbitrary estimates of the value-to-go and iteratively improve these estimates to eventually converge to the unknown optimal policy. We refer interested readers to \citet{gijsbrechts2018can} for an excellent review of ADP-based approaches for inventory management. Model-agnostic approaches circumvent the issue of partial or no knowledge of the underlying system dynamics by trial-and-error using an environment that generates immediate rewards and state transitions, given the current state and a selected action. This framework is popularly referred to as Reinforcement Learning (RL) \citep{sutton2018reinforcement}. RL methods themselves are categorized into model-based RL and model-free RL. In model-based RL, the transition functions are learned along with the optimal policy, while in model-free RL, the transition functions are not explicitly learned.

In this paper, we focus on model-free RL methods. One of the popular model-free RL algorithms is Q-learning, where the value of each state-action pair is estimated using the collected trajectory of states and rewards based on different actions, and the optimal action is obtained by exhaustive search. In classical RL methods, a set of features is chosen, and polynomial functions of those features are used to approximate the value function \citep{van1997neuro}. This approach initially achieved limited success. However, following the deep-learning revolution and the advent of significant computational power, deep-RL (DRL) methods experienced a resurgence in both popularity and effectiveness. This resurgence is attributed to the utilization of neural networks to approximate the value function, thereby automating the process of feature and function selection. This advancement facilitated numerous algorithmic breakthroughs, including the development of a family of policy gradient methods known as the actor-critic method \citep{mnih2016asynchronous}. In these methods, neural networks are employed to approximate both the value function and the policy itself. The policy is represented with an actor network that encodes a distribution over actions. Despite their successes, DRL and actor-critic approaches encounter several challenges, such as a lack of robust convergence properties, high sensitivity to hyperparameters, high sample complexity, and function approximation errors within their networks, leading to sub-optimal policies and incorrect value estimation \citep{pmlr-v80-haarnoja18b, maei2009convergent, fujimoto2018addressing, duan2016benchmarking, schulman2017proximal, henderson2018deep, lillicrap2016continuous}. To address these issues, new variations of the actor-critic approach continue to be proposed, such as Proximal Policy Optimization (PPO), which aims to avoid convergence to a sub-optimal solution while still enabling substantial policy improvement per update by constraining the divergence of the updated policy from the old one \citep{schulman2017proximal}, and Soft Actor-Critic (SAC), which improves exploration via entropy regularization to enhance hyperparameter robustness, prevent convergence to poor local optima, and accelerate policy learning \citep{pmlr-v80-haarnoja18b}.

The current work is complementary to this literature since we provide a principled way of factoring in known constraints and immediate rewards explicitly in training, as opposed to implicitly inferring or learning them. These aspects of our proposed approach potentially help address the known issues with DRL methods, such as reducing sample complexity, improving robustness, reducing the risk of convergence to poor solutions, and eliminating the dependence on function approximation of the policy network, which may be inaccurate due to overfitting or underfitting, or due to sampling issues from the data (e.g., under-exploration or sub-optimal convergence).

\paragraph{RL and Math Programming (MP) for inventory management:}
Early work demonstrating the benefits of RL for multi-echelon inventory management problems includes \citet{van1997neuro}, \citet{giannoccaro2002inventory}, and \citet{stockheim2003reinforcement}. Recently, there has been a surge in using DNN-based reinforcement learning techniques to solve supply chain problems \citep{de2022reward, gijsbrechts2018can, oroojlooyjadid2021deep, sultana2020reinforcement, hubbs2020or}. A DNN-based actor-critic method to solve the inventory management problem was studied by \citet{gijsbrechts2018can} for the cases of single node lost sales, dual sourcing settings, and multi-echelon settings, demonstrating improved performance in the latter. \citet{oroojlooyjadid2021deep} showed how RL could be used to solve the classical beer game problem where agents in a serial supply chain compete for limited supply. More recently, \citet{sultana2020reinforcement} used a multi-agent actor-critic framework to solve an inventory management problem for a large number of products in a multi-echelon setting. Similarly, \citet{qi2023practical} developed a practical end-to-end method for inventory management with deep learning. \citet{hubbs2020or} demonstrated the benefit of RL methods over static policies, like base stock, in a serial supply chain for a finite horizon problem. For an excellent overview and roadmap for using RL in inventory management, we refer interested readers to \citet{boute2021deep}. Unlike these studies, we adopt a mathematical programming-based RL actor and show its benefits over vanilla DRL approaches in the inventory management setting.

MP techniques have recently been used to optimize actions in RL settings with DNN-based function approximators and large action spaces. They leverage MP to optimize a mixed-integer (linear) problem (MIP) over a polyhedral action space using commercially available solvers such as CPLEX and Gurobi. Several studies have shown how trained ReLU-based DNNs can be expressed as an MP, with \citet{tjandraatmadja2020convex} and \citet{anderson2020strong} providing ideal reformulations that improve computational efficiency with a solver. \citet{ryu2019caql} proposed a Q-learning framework to optimize over continuous action spaces using a combination of MP and a DNN actor. \citet{delarue2020reinforcement}, \citet{van2019approximate}, and \citet{xu2020deep} demonstrated how to use ReLU-based DNN value functions to optimize combinatorial problems (e.g., vehicle routing) where the immediate rewards are deterministic and the action space is vast. In this paper, we apply these techniques to inventory management problems where in addition to the complexities mentioned above, the rewards are also uncertain.

	\section{Model and Performance Metrics}\label{sec:general_model}
 \paragraph{Model and notation:} We consider an infinite horizon discrete-time discounted Markov decision process (MDP) with the following representation: states $s \in \mathcal{S}$, actions $a \in \mathcal{A}(s)$, uncertain random variable $D \in \mathbb{R}^{\texttt{dim}}$ with probability distribution $P(D=d|s)$ that depends on the context state $s$, reward function $R(s, a, D)$, distribution over initial states $\beta$, discount factor $\gamma$, and transition dynamics $s' = \mathcal{T}(s, a, d)$, where $s'$ represents the next state. A stationary policy $\pi \in \Pi$ is specified as a distribution $\pi(\cdot|s)$ over actions $\mathcal{A}(s)$ taken at state $s$. The expected return of a policy $\pi \in \Pi$ is given by $J^{\pi} =  \mathbb{E}_{s \sim \beta} V^{\pi}(s)$, where the value function is defined as
\begin{center}
\( V^{\pi}(s) = \sum_{t=0}^\infty \mathbb{E}\left[\gamma^t R(s_t, a_t, D_t) \mid s_0 = s, \pi, P, \mathcal{T}\right]\,, \)
\end{center}
where the expectation is taken over both the immediate reward and the transition state. The optimal policy that maximizes the long-term expected discounted reward is given by $\pi^{*} := \arg \max_{\pi \in \Pi} J^{\pi}\,.$ Finally, by Bellman's principle, the optimal policy is the unique solution to the following recursive equation:
\begin{align}\label{eq:bellman}
V^{\pi^{*}}(s) = \max_{a \in \mathcal{A}(s)} \mathbb{E}_D\left[R(s, a, D) + \gamma V^{\pi^{*}}(\mathcal{T}(s, a, D))\right]\,.
\end{align}
As discussed in \S \ref{subsec:literature_review}, solving (\ref{eq:bellman}) directly is computationally intractable due to the curse of dimensionality. We take a hybrid approach where a model determines the immediate reward, but the value-to-go is model-free and is determined using trial-and-error in a simulation environment. We discuss the algorithm in detail next.

\subsection{Algorithm}\label{sec:general_PARL}
We propose a Monte Carlo simulation-based policy-iteration framework where the learned policy is the outcome of a mathematical program, which we refer to as PARL: Programming Actor Reinforcement Learning. As is common in RL-based methods, our framework assumes access to a simulation environment that generates state transitions and rewards, given an action and a current state. PARL is initialized with a random policy. The initial policy is iteratively improved over epochs with a learned critic (or the value function). In epoch \textit{j}, policy $\pi_{j-1}$ is used to generate \textit{N} sample paths, each of length \textit{T}. At each time step, a tuple of \{state ($s_t^n$), reward ($R_t^n$), next-state ($s_{t+1}^{n}$)\} is generated from the environment, which is then used to estimate the value-to-go function $\hat{V}^{\pi_{j-1}}_{\theta}$. The value-to-go function is represented using a neural network parameterized by $\theta$, which is obtained by solving the following error minimization problem:\\
\begin{center}
\vspace{-1em}
\(\min_{\theta}\sum_{n=1}^N\sum_{t=1}^T \left(R^{cum,n}_t - \hat{V}^{\pi_{j}}_{\theta}(s_t^n)\right)^2\,, \)
\end{center}
where the target variable $R^{cum,n}_t := \sum_{i=t}^T \gamma^{i-t} R^n_i$ is the cumulative discounted reward from the state at time $t$ in sample path $n$, generated by simulating policy $\pi_{j-1}$. Once $\hat{V}$ is estimated, the new policy using the trained value-to-go function is given by:
\begin{equation}\label{eq:per_step_problem}
\pi_{j}(s) = \arg\max_{a \in \mathcal{A}(s)} \mathbb{E}_D\left[R(s, a, D) + \gamma \hat{V}^{\pi_{j-1}}_{\theta}(\mathcal{T}(s, a, D))\right].
\end{equation}
Problem (\ref{eq:per_step_problem}) resembles Problem (\ref{eq:bellman}) except that the true value-to-go is replaced with an approximate value-to-go. Since each iteration leads to an updated and improved policy, we refer to it as a policy iteration approach. Next, we discuss how to solve Problem (\ref{eq:per_step_problem}) to obtain an updated policy in each iteration. Problem (\ref{eq:per_step_problem}) is challenging to solve for two main reasons. First, $\hat{V}^{\pi_{j-1}}$ is a neural network, making enumeration-based techniques intractable, especially in settings where the action space is large and combinatorial. Second, the objective function involves evaluating the expectation over the distribution of uncertainty \textit{D}, which could be analytically intractable to compute in many settings. We next discuss how PARL addresses each of these complexities.

\subsection{Optimizing over a neural network}

\begin{figure*}[h]
\centering
   \includegraphics[width=.6\textwidth]{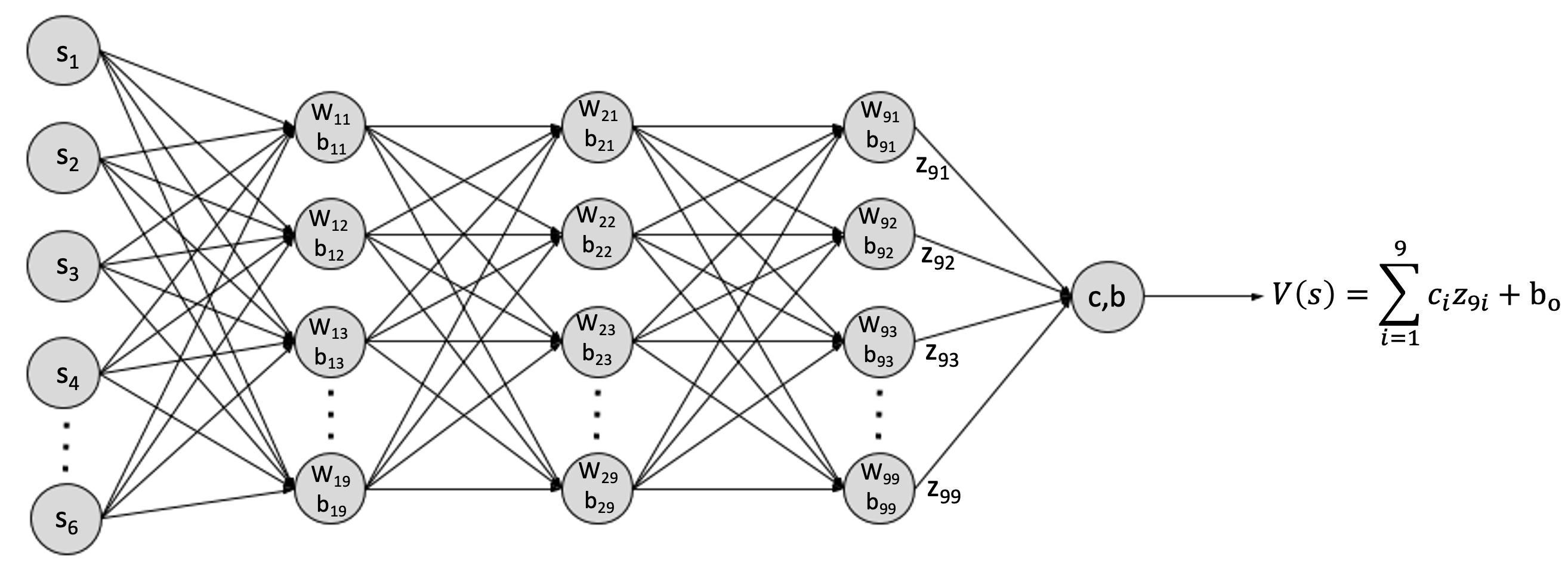}
         \caption{\small{A representative NN that takes as an input, a 6-dimensional state space, and considers a 10 layer NN (including the output layer) with each internal layer containing 9 neurons. Each neuron is defined by weights (W) and bias (b), except for the output layer that is defined with parameter vector $c$ and $b_o$. The output of each neuron uses ReLU activation and passes it as an input to the neurons in the subsequent layer.}}
         \label{fig:nn_figure}
         \vspace{-2em}
\end{figure*}

We first focus on the problem of maximizing the objective in (\ref{eq:per_step_problem}). To simplify the problem, we start by considering the case where demand $D$ is deterministically known to be $d$. Then, Problem (\ref{eq:per_step_problem}) can be written as:
\begin{center}
\(\max_{a \in \mathcal{A}(s)} R(s, a, d) + \gamma \hat{V}^{\pi_{j-1}}_{\theta} (\mathcal{T}(s, a, d))\,,\)
\end{center}
where we have removed the expectation over the uncertain demand $D$. Notice that the decision variable, $a$, is an input to the value-to-go estimate represented by $\hat{V}$. Hence, optimizing over $a$ involves optimizing over a neural network, which is non-trivial. We take a mathematical programming-based approach to solve this problem. First, we assume that the value-to-go function is a trained $K$-layer feed-forward ReLU network with input state $s$ and satisfies the following equations for all $k = 2, \ldots, K$:
\begin{center}
\(
z_1 = s, ~\hat{z}_k = W_{k-1} z_{k-1} + b_{k-1}, ~
z_k = \max\{0, \hat{z}_k\}, ~
\hat{V}_{\theta}(s) := c^T \hat{z}_K +b_o\,.
\)
\end{center}
Here, $\theta = (c, \{(W_k, b_k)\}_{k=1}^{K-1}+b_o$ are the parameters of the value-to-go estimator. Specifically, $(W_k, b_k)$ are the multiplicative and bias weights of layer $k$, and $c$ and $b_o$ are the weights and the bias of the output layer (see Figure \ref{fig:nn_figure}).

Finally, $\hat{z}_k$ and $z_k$ denote the pre- and post-activation values at layer $k$. ReLU activation at each neuron allows for a concise mathematical programming representation that uses binary variables and big-M constraints \citep{ryu2019caql, anderson2020strong}. For completeness, we briefly describe the steps next. Consider a neuron in the neural network with parameters $(w, b)$. For example, in layer $k$, neuron $i$'s parameters are $(W_k^i, b_k^i)$. Assuming a bounded input $x \in [l, u]$, the output $z$ from that neuron can be obtained by solving the following MP representation:
\begin{equation}
\begin{aligned}
 \max_{z, y}~~ 0 ~ s.t.~ \{(z,y)|
z \geq w^T x + b,
~z \geq 0,
~z \leq w^T x + b - M^-(1-y),
~z \leq M^+ y, ~y \in \{0, 1\}\,.\}
\end{aligned}
\label{eq:NN_representation}
\end{equation}
We refer to the constraint set concisely as \texttt{P}($w$, $b$, $x$). Here, $M^+ = \max_{x \in [l, u]} (w^T x + b)$ and $M^- = \min_{x \in [l, u]} (w^T x + b)$ are the maximum and minimum outputs of the neuron for any feasible input $x$. Note that $M^+$ and $M^-$ can be easily calculated by analyzing the component-wise signs of $w$. For example, let
\begin{center}
\(
\tilde{u}_i = \begin{cases}
u_i & \textit{if}~ w_i \geq 0, \\
l_i & \text{otherwise},
\end{cases}
\quad \& \quad
\tilde{l}_i = \begin{cases}
l_i & \textit{if}~ w_i \geq 0, \\
u_i & \text{otherwise}.
\end{cases}
\)
\end{center}
Then, simple algebra yields that $M^+ = w^T \tilde{u} + b$ and $M^- = w^T \tilde{l} + b$. Starting with the bounded input state $s$, the upper and lower bounds for subsequent layers can be obtained by assembling the $\max\{0, M^+\}$ and $\max\{0, M^-\}$ for each neuron from its prior layer. We will refer to them as $[l_k, u_k]$ for every layer $k$. This MP reformulation of the neural network that estimates the value-to-go is crucial in our approach. Particularly, since we can also represent the immediate reward directly in terms of the decision $a$, and the feasible action set for any state is a polyhedron, we can now use the machinery of integer programming to solve Problem (\ref{eq:per_step_problem}). We will make this connection more precise in \S \ref{sec:inventory} in the context of inventory management.

Next, we discuss how to tackle the problem of estimating the expectation in Problem (\ref{eq:per_step_problem}).

\subsection{Maximizing expected reward with a large action space}

The objective in Problem (\ref{eq:per_step_problem}) involves an expectation taken over the uncertainty \( D \). Note that the uncertainty in \( D \) impacts both the immediate reward as well as the value-to-go via the transition function. Evaluating this expectation could be potentially difficult since \( D \) generally has a continuous distribution and we use a neural network (NN) based value-to-go approximator. Hence, we take a Sample Average Approximation (SAA) approach \citep{kim2015guide} to solve it. Let \( d_1, d_2, \ldots, d_{\eta} \) denote \( \eta \) independent realizations of the uncertainty \( D \). Then, SAA approximates the expectation as follows:
\begin{center}
\(\mathbb{E}_{D}[R(s, a, D) + \gamma \hat{V}^{\pi_{j-1}}_{\theta}(\mathcal{T}(s, a, D))] \approx \frac{1}{\eta}\sum_{i=1}^{\eta} \left(R(s, a, d_i) + \gamma \hat{V}^{\pi_{j-1}}_{\theta}(\mathcal{T}(s, a, d_i))\right)\,.\)
\end{center}
Hence, Problem (\ref{eq:per_step_problem}) becomes:
\begin{equation}\label{eq:SAA}
\hat{\pi}^{\eta}_{j}(s) = \arg \max_{a \in \mathcal{A}(s)} \frac{1}{\eta} \sum_{i=1}^{\eta} \left(R(s, a, d_i) + \gamma \hat{V}^{\pi_{j-1}}_{\theta}(\mathcal{T}(s, a, d_i))\right).
\end{equation}

Problem (\ref{eq:SAA}) involves evaluating the objective only at sampled points instead of all possible realizations. Assuming that for any \( \eta \), the set of optimal actions is non-empty, we show that as the number of samples \( \eta \) grows, the estimated optimal action converges to the optimal action. We make this statement precise in Proposition \ref{prop:SAA_guarantee}. The proof follows through standard results in the analysis of SAA and is provided in Appendix \S\ref{app:SAA_guarantee}.

\begin{proposition}\label{prop:SAA_guarantee}
Consider epoch \( j \) of the PARL algorithm with a ReLU-network value function estimate \(\hat{V}_{\theta}^{\pi_{j-1}}(s)\) for some fixed policy \(\pi_{j-1}\). Suppose \(\pi_j\) and \(\hat{\pi}^{\eta}_j\) are the optimal policies as described in Problem (\ref{eq:per_step_problem}) and its corresponding SAA approximation, respectively. Then, for all \( s \), \(\lim_{\eta \to \infty} \hat{\pi}^{\eta}_j(s) = \pi_j(s).
\)
\end{proposition}

We are now in a position to discuss the proposed PARL algorithm. (see Algorithm \ref{alg:PARL} and the illustrative block diagram in Figure \ref{fig:PARL} of Appendix \ref{app:notation_and_block}).

\begin{algorithm}[h]
\footnotesize
\begin{algorithmic}[1]
\State Initialize with random policy $\pi_0$, number of epochs $\mathcal{J}$, number of episodes $N$, number of steps $T$ and exploration rate $\epsilon$.
\For {(epoch) $j \in[\mathcal{J}]$}
    \For {(episode) $n \in [N]$}
        \State Play policy $\pi_{j-1}$ for $T(1-\epsilon)$ steps and random actions for $\epsilon T$ steps, starting with state  $s_0^n \sim \beta$.
        \State Let $R^{cum,n}_t = \sum_{i=t}^{T} \gamma^{i-t} R_{i}^n$ and store tuple \{$s_t^n$, $R_t^{cum,n}$\}~$\forall t=1,\ldots,T$.
    \EndFor
    \State Approximate a DNN value function using backward induction by solving the following mean squared error minimization problem:
    \begin{center}
    \(\hat{V}_{j} = \arg \min_{\theta} \sum_{n=1}^N \sum_{t=1}^T \left(R^{cum,n}_t - f(s_t^n, \theta)\right)^2\)
    \end{center}
    \State Sample $\eta$ realizations of the underlying uncertainty $D$ and obtain a new policy (as a lazy evaluation, as needed) by solving Problem (\ref{eq:SAA}).
\EndFor
\end{algorithmic}
\caption{PARL}
\label{alg:PARL}
\end{algorithm}
For each of the $\mathcal{J}$ epochs (outer loop iterations) (line 2), we run the current policy (compute actions using our optimization approach and current value network) for some number of steps.  Specifically we run it for $N$ random episodes (line 3), each consisting of $T$ steps (line 4) - to collect a batch of trajectories consisting of action and (state, reward) tuples.  Note in each episode (trajectory), we compute approximate discounted cumulative rewards per state in the trajectory according to the definition, using the discount factor and the collection of observed future rewards from that state (line 5). This provides examples of state and discounted cumulative rewards collected, which serve as inputs and outputs, respectively, for the value neural network - a neural net to predict the value (discounted cumulative reward) given the state, so it can be trained and updated using these examples in the standard way. We update the parameters of the value network by training it with standard stochastic gradient descent approaches for some number of update epochs (neural net training epochs) - using this batch of data collected (line 7).  This is the standard neural net training - where in each outer loop epoch ($j$) we update the current parameters of the network - i.e., they are randomly initialized at the beginning and updated with each new batch of trajectories from the current policy.  Finally, (8) defines the updated policy which results from our stochastic optimization approach given the current value network for a given current state input.  Note we only compute this when needed (lazily) - when specifically evaluating the policy for a particular state.  Also note, this policy is used in line (4) implicitly - to play the current policy - i.e., for each state we must use this current policy to choose the next action. Finally, as is commonly done in RL algorithms, we use $\epsilon$-greedy approach \citep{sutton2018reinforcement} to ensure exploration while training the value-network.\footnote{Proposition 1 shows the convergence of the estimated policy in each epoch as we scale the number of demand samples. Indeed to demonstrate convergence to the optimal policy in the finite samples case of PARL, one has to account for two sources of error: (i) value estimation error on account of using DNNs to approximate the value function with finite reward trajectories and (ii) the finite number of demand samples to compute the state transition, and in-turn the \textit{expected} immediate reward as well as future value-to-go estimates for different actions. Nevertheless, one can leverage recent results on the sample complexity analysis of Neural Networks \cite{yarotsky2017error,golowich2018size,bartlett2019nearly} along with the analysis of Approximate Policy Iteration (API) and Approximate Value Iteration (AVI) \cite{bertsekas1996neuro,munos2003error,munos2008finite} to show convergence of the estimated policy to the optimal policy even with finite samples. }

	\section{Inventory Management Application}\label{sec:inventory}
We now describe the application of PARL to an inventory management problem.\footnote{We note that our proposed framework is useful in settings where (i) the underlying decision problem can be modeled as a Markov Decision Process (MDP), where the decision maker has information on the evolution of the states, rewards, and the transition function; and (ii) the action space is large (combinatorial), making the decision problem difficult to solve exactly. Apart from the inventory replenishment problem discussed in the following sections, another setting where our proposed framework might be applicable is Network Revenue Management (NRM). NRM problems primarily involve a decision maker with limited quantities of multiple resources who sells products, each composed of a bundle of resources. The decision aims to maximize revenue by deciding the availability or prices of products over time \citep{farias2007approximate}. A classical example of NRM is pricing fares in the airline industry. Airlines operate flights on a network of locations and have to price different origin-destination pairs on this network over time. In this problem, demand is stochastic, there is limited capacity on each leg, and the action space is large. NRM is often formulated as a dynamic programming problem, which is typically computationally intractable. Hence, various heuristics have been proposed to solve this problem, such as fare class, bid-price heuristics, and other approximate dynamic programming heuristics \citep{topaloglu2009using,zhang2009approximate}} We consider a firm managing inventory replenishment and distribution decisions for a single product across a network of stores (also referred to as nodes) with the goal of maximizing profits while meeting customer demands. Our objective is to incorporate practical considerations in the inventory replenishment problem, specifically accounting for (i) lead times in shipments, (ii) fixed and variable costs of shipment, and (iii) lost sales due to unfulfilled demand. These problems are notoriously difficult to solve, and there are few results on the structure of the optimal policy (see \S \ref{subsec:literature_review}). However, as we will describe, RL, and particularly PARL, can be used to generate well-performing policies for these problems.

\subsection{Model and System Dynamics}
The firm's supply chain network consists of a set of nodes ($\Lambda$), indexed by \( l \). These nodes can represent warehouses, distribution centers, or retail stores. Each node in the supply chain network can produce inventory units (denoted by the random variable \( D^{p}_l \)) and/or generate demand (denoted by the random variable \( D^{d}_l \)). Inventory produced at each node can be stored at the node or shipped to other nodes in the supply chain. For node \( l \), \( O_l \) denotes the upstream nodes that can ship to node \( l \). At any given time, each node can fulfill demand based on the available on-hand inventory at the node. We assume that any unfulfilled demand is lost (the \textit{lost-sales} setting).

The firm's objective is to maximize revenue (or minimize costs) by optimizing different fulfillment decisions across the nodes of the supply chain. Fulfillment at any node can occur both through transshipment between nodes in the supply chain and from an external supplier. Each transshipment from node \( l \) to \( l' \) has a deterministic lead time \( L_{ll'} \geq 0 \) and is associated with a fixed cost \( K_{ll'} \) and a variable cost \( C_{ll'} \). The fixed cost could be related to hiring trucks for the shipment, while the variable cost could be related to the physical distance between the nodes. Each node \( l \) has a holding cost of \( h_l \). Each inventory unit sold at different nodes generates a profit (equivalently revenue) of \( p_l \). Next, we discuss the system dynamics in detail next. Note that the dependence on the time period \( t \) is suppressed for ease of exposition.

\begin{enumerate}
    \item In each period \( t \), the firm observes \(\mathbf{I}\), the inventory pipeline vector of all nodes in the supply chain. The inventory pipeline vector for node \( l \) stores the on-hand inventory as well as the inventory that will arrive from upstream nodes. By convention, we denote \( I^0_l \) as the on-hand inventory at node \( l \).
    \item The firm makes trans-shipment decision \( x_{l'l} \)\footnote{As is conventionally done in inventory management problems, we denote actions with the variable \( x \) instead of \( a \) in this and the following sections.}, which denotes the inventory to be shipped from node \( l' \) to \( l \) and incurs a trans-shipment cost (\texttt{tsc}) for node \( l \):
  \(
    \mathtt{tsc}_l = \sum_{l' \in O_{l}} \left[ K_{l' l}\mathbbm{1}_{x_{l'l} > 0} +  C_{l'l} x_{l'l} \right]\,.
    \)
    \item The available on-hand inventory at each node is updated to account for units that are shipped out, units that arrive from other nodes, and units that are produced at this node. We let \( \tilde{I}_l^0 \) denote this intermediate on-hand inventory, which is given by:
    \begin{center} \(\tilde{I}_l^0 = I_{l}^0 + I_{l}^{1} + D^p_{l} + \sum_{l' \in O_l} x_{l'l}\mathbbm{1}_{L_{l'l}=0} - \sum_{\{l' \in \Lambda \mid l \in O_{l'}\}} x_{ll'}\,.
    \)\end{center}
    \item The stochastic demand at each node, \( D^d_l \), is realized and the firm fulfills demand from the intermediate on-hand inventory, generating a revenue from sales (\texttt{rs}) of \( \mathtt{rs}_l = p_l \min\{D^d_l, \tilde{I}_l^0\} \).
    \item Excess inventory (over capacity) gets salvaged, and the firm incurs holding costs on the leftover inventory. The firm incurs holding-and-salvage costs (\texttt{hsc}) given by:
    \begin{center}
    \(\mathtt{hsc}_l = h_l \min \left\{\bar{U}_l, \left[ \tilde{I}^0_l - D^d_{l}\right]^+\right\} + \delta_l [\tilde{I}^0_l - D^d_{l} - \bar{U}_l]^+\,,\)
    \end{center}
    where \( \bar{U}_l \) is the storage capacity of node \( l \). We let \( I_{l}^{\prime 0} := \min \left\{\bar{U}_l, \left[ \tilde{I}^0_l - D^d_{l}\right]^+\right\} \) for ease of notation.
    \item Finally, inventory gets rotated at the end of the time period:
    \begin{center}
    \(I_{l}^{\prime j} = I_{l}^{j+1} + \sum_{l' \in O_l} x_{l'l} \mathbbm{1}_{L_{l'l}=j}, \quad \forall\ 1 \leq j \leq \max_{l' \in O_{l}} L_{l'l}\,.\)
    \end{center}
    The rotated inventory becomes the inventory pipeline for the next time period. That is, \( \mathbf{I}_{t+1} = I^{\prime}_t \).
\end{enumerate}

The problem of maximizing profits in this replenishment problem can now be formulated as an MDP. In particular, the state space \( s \) is the pipeline inventory vector \( \mathbf{I} \); the action space is defined by the set of feasible actions; the transition function \( \mathcal{T} \) is defined by the set of next-state equations (which depend on the distribution of demand \( D^d_l \)); and the reward from each state is the profit minus the inventory holding and trans-shipment costs in each period. We can write the optimization problem using Bellman recursion. Let
\(\! R_l(\mathbf{I}, \mathbf{x}, D) = \mathtt{rs}_l - \mathtt{tsc}_l - \mathtt{hsc}_l \!\)
denote the revenue per node as a function of the pipeline inventory, the trans-shipment decisions, and the stochastic demand and production. Then, the total revenue generated from the supply chain in each time period is
\begin{center}
\(\overset{*}{R}(\mathbf{I}, \mathbf{x}, D) = \sum_{l \in \Lambda} R_l(\mathbf{I}, \mathbf{x}, D).\)
\end{center}
Similarly, the optimization problem can be written as
\begin{center}
\(V(\mathbf{I}) = \max_{\mathbf{x} \in \mathcal{A}(\mathbf{I})} \mathbb{E}_{D} \left[ \overset{*}{R}(\mathbf{I}, \mathbf{x}, D) + \gamma V(\mathbf{I}^{\prime}) \right],\)
\end{center}
where \( \mathbf{I}^{\prime} \) is implicitly a function of \( \mathbf{I}, \mathbf{x} \), and \( D \). As discussed before, this inventory replenishment problem takes exactly the same form as the general problem in \S \ref{sec:general_model}. It can be solved using Bellman recursion, which is unfortunately computationally intractable. Hence, in what follows, we will discuss how the framework developed in \S \ref{sec:general_PARL} can be effectively used to solve such problems.

\subsection{PARL for Inventory Management}\label{sec:MPforPARL}
Recall that the PARL algorithm solves Problem (\ref{eq:per_step_problem}) to estimate an approximate optimal policy. In the inventory management context, this problem becomes
\begin{center}
\(\pi(\mathbf{I}) = \argmax_{\mathbf{x} \in \mathcal{A}(\mathbf{I})} \frac{1}{\eta} \sum_{i=1}^{\eta} \overset{*}{R}(\mathbf{I}, \mathbf{x}, d_i) + \gamma \hat{V}(\mathbf{I}^{\prime}),\)
\end{center}
where, as discussed before, the next state $\mathbf{I}^{\prime}$ is a function of the current state, action, and the demand realization. We describe how this problem can be written as an integer program. First, since we consider the lost-sales inventory setting, we define auxiliary variables $\texttt{sa}_{li}$ that denote the number of units sold for node $l$ and demand sample $i$. Then, the constraints \(\! \texttt{sa}_{li} \leq d^d_{li} ~\&~
    \texttt{sa}_{li} \leq \tilde{I}_{li}^0,\!\) ensure that the number of units sold is less than or equal to the inventory on-hand and demand. Note that with discounted rewards and time-invariant prices/costs, opportunities for stock hedging in future time periods due to the presence of a sales variable are not present, and that sales will exactly be the minimum of demand and inventory. We also define auxiliary variables $B_{li}$ that denote the number of units salvaged at node $l$ for demand sample $i$. Then, the constraints
\[
\begin{split}
    \tilde{I}_{li}^{0} &= I_{l}^0 + I_{l}^{1} + d^p_{li} + \sum_{l' \in O_l} x_{l'l} \mathbbm{1}_{L_{l'l}=0} - \sum_{\{l' \in \Lambda \mid l \in O_{l'}\}} x_{ll'}, \\
    I^{'0}_{li} &= \tilde{I}^{0}_{li} - \texttt{sa}_{li} - B_{li}, \\
    I^{'j}_{li} &= I_{l}^{j+1} + \sum_{l' \in O_l} x_{l'l} \mathbbm{1}_{L_{l'l}=j},~ \forall 1 \leq j \leq \max_{l' \in O_l} L_{l'l},
    ~B_{li} \geq 0,
\end{split}
\]
capture the next state transition for each demand realization. Finally, the objective function has two components: the immediate reward and the value-to-go. The immediate reward, in terms of the auxiliary variables, can be written as
\begin{center}
\(R_{l}(\mathbf{I}, \mathbf{x}, d_i) = \underbrace{p_{l} \texttt{sa}_{li}}_{\texttt{rs}_l} - \underbrace{\sum_{l^{'} \in O_l} \left[ K_{l^{'} l} \mathbbm{1}_{x_{l'l} > 0} + C_{l^{'}l} x_{l^{'}l} \right]}_{\texttt{tsc}_l} - \underbrace{(h_{l} I^{'0}_{li} + \delta B_{li})}_{\texttt{hsc}_l}.\)
\end{center}

Note that $g_{l'l}$ is a binary variable that models the fixed cost of ordering. Next, let $\mathbf{I}^{\prime}_i$ denote the next state under the $i^{th}$ demand realization. Assume that the NN estimator for the value-to-go has $\Psi$ fully connected layers with $\mathcal{N}_k$ neurons in each layer with ReLU activation and an output layer. Let $\texttt{neu}_{jk}$ denote the $j^{th}$ neuron in layer $k$. Then, the outcome from the neurons in the first layer with the $i^{th}$ demand realization can be represented as
\begin{center}
\((z^i_{j1}, y^i_{j1}) \in \texttt{P}(W_{1}^{j}, b_{1}^{j}, \mathbf{I}^{\prime}_i), \quad \forall j \in [1, \ldots, \mathcal{N}_1]\,,\)
\end{center}
where recall that $\texttt{P}(\cdot, \cdot, \cdot)$ defined in Eq. (\ref{eq:NN_representation}) denotes the MP constraint set used to represent the output from the respective neuron. The output from this layer becomes the input of the next layer. Hence, let $\textbf{Z}^i_1 := [z_{11}, z_{21}, \ldots, z_{\mathcal{N}_{1}1}]$ denote the outcome of layer 1. Then, the output of each neuron of layer 2 with the $i^{th}$ demand realization can be written as
\begin{center}
\((z^i_{j2}, y^i_{j2}) \in \texttt{P}(W_{2}^j, b_{2}^j, \textbf{Z}^i_1),~ \forall j \in [1, \ldots, \mathcal{N}_2]\,.\)
\end{center}
Continuing this iterative calculation, we have that $\forall k = 2, \ldots, \Psi$
\begin{center}
\((z^i_{jk}, y^i_{jk}) \in \texttt{P}(W_{k}^j, b_{k}^j, \textbf{Z}^i_{k-1}), \quad \forall j \in [1, \ldots, \mathcal{N}_k]\,.\)
\end{center}
Finally, letting $c, b_o$ denote the weight vector and bias of the output layer, we have that $V(\mathbf{I}^{\prime}_i) = c^\top \textbf{Z}_{\Psi} + b_o,$
where the value-to-go is implicitly a function of the decisions $x$ since they impact the next state $\mathbf{I}^{\prime}$. Using this value-function approximation, the inventory fulfillment problem for each time period can now be written as
\allowdisplaybreaks
\begin{subequations}
\begin{align}
\max_{ x_{l'l}\in \mathbb{Z}^+, {\bf U^L } \leq {\bf x} \leq {\bf U^H }}
                  & \frac{1}{\eta} \sum_{i=1}^{\eta} \left[ \overset{*}{R}({\bf I}, {\bf x}, {\bf d}_i) + \gamma \left(c^T {\bf Z}^{i}_{\Psi} + b_0\right) \right]\\
\text{where} \quad  \overset{*}{R}({\bf I}, {\bf x}, {\bf d}_i)  &= \sum_{l \in \Lambda}  R_{l}( {\bf I}_{l}, {\bf  x}_{l}, {\bf d}_{li}) \\
   R_{l}({\bf I}_{l}, {\bf x}_{l}, {\bf d}_{li}) &=  \texttt{rs}_l - \texttt{tsc}_l - \texttt{hsc}_l \\
 \label{saleslessdemand} \texttt{sa}_{li} &\leq  d^{d}_{li}, ~\texttt{sa}_{li} \leq \tilde{I}_{li}^{0} \quad \forall l \in \Lambda, i, \\
 \label{indicator1} g_{ll^{'}} &\leq x_{ll^{'}}, ~x_{ll^{'}} \leq U^H_{ll^{'}} g_{ll^{'}} \quad \forall l \in O_{l^{'}}, l' \in \Lambda, \\
  \tilde{I}_{li}^{0} &=  I_{l}^{0} + I_{l}^{1} + d^{p}_{li} + \sum_{l' \in O_l} x_{l'l} \mathbbm{1}_{L_{l'l}=0} - \sum_{\{l' \in \Lambda \mid l \in O_{l'}\}} x_{ll'}, \quad \forall l \in \Lambda, i, \\
   I^{'0}_{li} &=  \tilde{I}^{0_{li}} - \texttt{sa}_{li} - B_{li}^{0}, \quad \forall l \in \Lambda, i, \\
   I^{'j}_{li} &= I_{l}^{j+1} + \sum_{l^{'} \in O_l} x_{l^{'}l} \mathbbm{1}_{L_{l^{'}l}=j} - B^{j}_{li}, \quad \forall 1 \leq j \leq \max_{l^{'} \in O_l} L_{l^{'}l}, l \in \Lambda, i, \\
 B_{li}^{j} & \geq 0, \quad \forall j = 0, \cdots, \max_{l' \in O_l} L_{l'l}, l \in \Lambda, i, \\
  (z^i_{j1}, y^i_{j1}) &\in \texttt{P}(W_{1}^{j}, b_{1}^{j}, \mathbf{I}^{\prime}_i), \quad \forall j \in [1, \ldots, \mathcal{N}_1], \forall i \in [1, \ldots \eta]\\
  (z^i_{jk}, y^i_{jk}) &\in \texttt{P}(W_{k}^j, b_{k}^j, \textbf{Z}^i_{k-1}), \quad \forall j \in [1, \ldots, \mathcal{N}_k],  \forall k \in [2, \cdots, \Psi], \forall i \in [1, \ldots \eta].
\end{align}
\end{subequations}
This problem is solved in each time-step (Step 8 of PARL, see Algorithm \ref{alg:PARL}) to generate reward and action trajectories that are in turn used to update and improve learned policies. Finally, for ease of understanding, in Appendix \ref{app:simplified_inv_example}, we further discuss this IP formulation in a simplified inventory replenishment example with a small supply chain network and a representative NN approximator.

\remark[Alternate Q-Learning Based Approach]
We note that the proposed PARL methodology is based on using the value function estimate $\hat{V}(s)$ to optimize actions. Alternatively, one could consider solving the following problem:
\begin{equation}\label{eq:Q_learning}
\pi_{j}(s) = \arg\max_{a \in \mathcal{A}(s)} \hat{Q}^{\pi_{j-1}}(s,a)\,,
\end{equation}
where $\hat{Q}^{\pi_{j-1}}(s,a)$ is the estimated expected discounted reward from selecting action $a$ when in state $s$. We decided not to pursue the Q-learning approach for multiple reasons. First and foremost, the larger size of the neural network needed to estimate the Q values is a significant drawback. 
Assume that the supply chain network is fully connected and that in each time period, the centralized planner decides on the number of units to be shipped from each supply chain node to another. Then, the action decision $a \in \mathbb{R}^{|\Lambda|^2}$. Hence, the input layer of the neural network approximating the Q-function will be of size $|\mathbf{I}| + |\Lambda|^2$. In comparison, the input layer in the value function-based approach we propose is of size $|\mathbf{I}|$, which is much smaller. Note that the larger the input space, the larger the DNN network size used for approximation. Since the number of integer variables increases linearly with the network size, smaller networks are preferred to ensure computational feasibility. Second, in a Q-learning based approach, directly modeling the immediate reward information is infeasible unless one solves two optimization problems per action step.

\section{Numerical Results and Managerial Insights  }\label{subsec:numerical_results}

In this section, we present numerical results on the performance of the proposed PARL algorithm in various supply chain settings. The objective is two-fold: (i) to benchmark the proposed algorithm and compare its performance with state-of-the-art RL and inventory management policies, and (ii) to discuss the usage of an open-source Python library to create inventory management simulation environments with the implementation of various RL algorithms, for easy benchmarking in supply chain applications. We model a representative supply chain network in this environment with at most three different node types:
\begin{itemize}
\item \textit{Supplier:} Supplier nodes (denoted by \texttt{S}) in the network produce (or alternatively order) inventory to be distributed across the network. For example, these can be large manufacturing sites that produce inventory units or a port where inventory from another country lands.
\item \textit{Retailer:} Retailer nodes (denoted by \texttt{R}) in the network consume inventory by generating demand for the product. For example, these can be retail stores that directly serve customers who are interested in purchasing the inventory units.
\item \textit{Warehouse:} Warehouse nodes (denoted by \texttt{W}) are intermediate nodes that connect supplier nodes to retailer nodes. They hold inventory and ship it to downstream retail nodes.
\end{itemize}
Each of the nodes is associated with holding costs, holding capacities, and spillage costs, while retailers are additionally associated with price, demand uncertainties, and a lost-sales/backorder demand type, and suppliers with production uncertainties. The directed links between the nodes form the supply chain network. Each link is associated with order costs, lead time, and maximum order quantity. The environment executes the ordering and distribution actions specified by the {\em policy} by first ensuring their feasibility using a proportional fulfillment scheme (as it cannot send more than the inventory in a node), samples the uncertainties, accumulates the {\em reward} (the revenue from fulfillment less the cost of ordering and holding), and returns the {\em next state}. With this overview, we start by first describing the various supply chain networks that we will consider for the numerical study.

	\subsection{Network structure and settings used for numerical experiments}
		\begin{figure}[t]
\centering
\includegraphics[width=.6\linewidth]{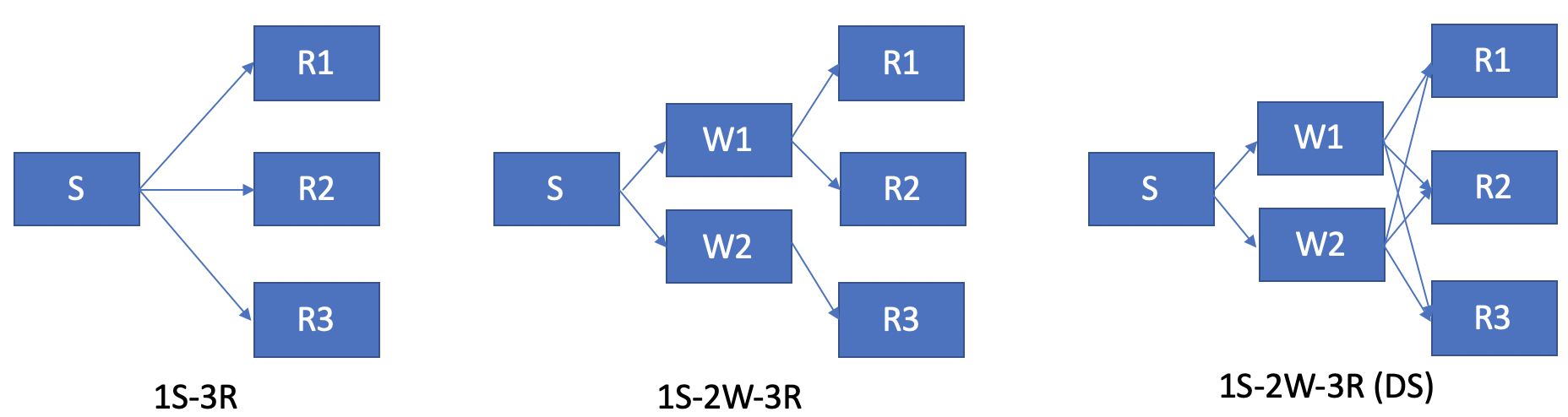}
\caption{Example of different multi-echelon supply chain networks. In 1S-3R, a single supplier node serves a set of 3 retail nodes directly. In 1S-2W-3R, the supplier node serves the retail nodes through  two warehouses. In 1S-2W-3R (dual sourcing), each retail nodes can is served by two distributors.}\label{SCexamples}
\vspace{-2em}
\end{figure}

We consider three different multi-echelon supply chain network structures, inspired by real-world retail distribution networks.
\begin{enumerate}
    \item \textit{Two-echelon networks:} This network consists of one supplier connected to a set of heterogeneous retailers that differ in terms of holding costs and lead time to ship from the supplier to the retailer. We consider networks with varying numbers of retailers (3, 10, and 20 retailers), as well as a setting with high versus low production. This is done to test the performance of the algorithms in resource-constrained settings where the inventory units available to be shipped could be less than the demand generated across the network. Hence, the policy must decide where to ship available inventory units. This setting is particularly inspired by fast-fashion retail, where the number of units per stock keeping unit (SKU) is very low. These settings are henceforth referred to as the 1\texttt{S}-m\texttt{R} networks. We specifically study the 1\texttt{S}-3\texttt{R}, 1\texttt{S}-10\texttt{R}, and 1\texttt{S}-20\texttt{R} in the low production setting and 1\texttt{S}-3\texttt{R}-High for the relatively higher production setting.
    \item \textit{Tree-distribution networks:} We study distribution networks with a tree structure, specifically networks consisting of one supplier, two intermediate warehouses, and three retailers that are each served from one of the two warehouses. This models networks where retailers own large warehouses that are geographically dispersed, with each serving a set of nearby retail stores. We analyze two different settings: 1\texttt{S}-2\texttt{W}-3\texttt{R}, where the supplier inventory is constrained, and 1$\texttt{S}^{\infty}$-2\texttt{W}-3\texttt{R}, which is the more traditional infinite inventory setting where the supplier is not constrained by inventory. Note that only when the supplier inventory is infinite, denoted by $\texttt{S}^{\infty}$, we do not include its inventory level as part of the state space.
    \item \textit{Distribution networks with dual-sourcing:} This setting is similar to the previous setting, except that the retail store is connected to multiple warehouses, specifically two. It models multi-echelon supply chains where retailers have large warehouses, sometimes farther away from demand centers with longer lead times, and smaller regional warehouses that also serve the demand with shorter lead times but relatively higher costs. Hence, the retailer must decide not only how much inventory to store, but also from which warehouse to ship this inventory. We refer to this setting as 1\texttt{S}-2\texttt{W}-3\texttt{R} (DS).
\end{enumerate}

In Table \ref{table:SCsettings} of Appendix \ref{app:parameter_table}, we present the different parameters of the supply chain in each setting. We discuss some salient features of these settings below. (i) \textit{Highly uncertain demand:} In many real-world settings (e.g., fast-fashion retail), demand for products is highly uncertain. Hence, we let the demand distribution have a low signal-to-noise ratio. (ii) \textit{Non-symmetric retailers:} The retailer node parameters are selected so that they differ in terms of (i) holding costs and (ii) lead time from the upstream warehouse. This asymmetry implies that commonly used heuristics, such as learning an approximate policy for one retailer and identically applying it to other retailers, would be highly sub-optimal. (iii) \textit{Fixed ordering costs, holding capacities, non-zero lead times, and lost sales:} All these settings are very common in practice. For example, in retail B2C distribution networks, retailers lose sales opportunities if the item is not on the shelf. There is also limited shelf space compared to upstream warehouses. Furthermore, these are settings where optimal policy structures have not been characterized and analytical tractability is not guaranteed.

\subsection{Benchmarks }

We compare PARL with four state-of-the-art, widely used RL algorithms: PPO \citep{schulman2017proximal}, TD3 \citep{fujimoto2018addressing}, SAC \citep{pmlr-v80-haarnoja18b}, and A2C \citep{mnih2016asynchronous}. For the RL algorithms, we used the tested and reliable implementations provided by Stable-Baselines3 \citep{stable-baselines3}, under the MIT License. We made all our environments compatible with OpenAI Gym \citep{brockman2016}, and to implement PARL, we built on reference implementations of PPO provided in SpinningUp \citep{SpinningUp2018} (both MIT License). We ran RL baselines on a 152-node cluster with an average of 26 CPUs per node (individual jobs used 1 CPU and less than 1GB RAM), and PARL on a 13-node cluster with an average of 48 CPUs per node (individual PARL jobs use 16 CPUs for trajectory parallelization and CPLEX computations and average less than 4GB RAM). We use version 12.10 of CPLEX with a time constraint of 60s per decision step with 2 threads.

We additionally compare PARL with commonly used order-up-to heuristics in supply chains: a popularly used $(s,S)$ base stock (BS) policy \citep{scarf1960optimality}, here implemented for each link (see Appendix~\ref{sec:basestock_heuristic} for implementation details); and a decomposition-aggregation (DA) heuristic \citep{rong2017heuristics} to evaluate echelon order-up-to policies $(S,S)$ for a serial path in near closed form (see Appendix~\ref{sec:DA_heuristic} for implementation details). The latter heuristic is designed for tree-networks with backordered demand and no fixed costs in the presence of infinite supply, and it has near-closed form expressions with asymptotic guarantees under those settings. In the former heuristic, a simulation-based grid search using the environment is performed to identify the best $(s,S)$ pair for every link, assuming infinite supply but with all other complexities (lost sales and fixed costs) intact. In fact, due to these differences, we believe that the BS heuristic tends to outperform the DA heuristic in the settings we study, as we will see in the next section.

\subsection{Testing Framework}
Evaluating RL algorithms can be tricky as most methods have a large number of hyperparameters or configurations that could be tuned. Care must also be taken to capture the actual utility of an RL method via the evaluation and avoid misleading conclusions due to improper evaluation \citep{henderson2018deep, agarwal2021deep}. In particular, since the same simulation environments are used to train and evaluate (test) a model, if a model is chosen based on its performance on that environment during a particular training run and the exact same model (model weights) is then used to evaluate on the same environment for test scoring, this can lead to misleading conclusions. The particular model class and set of hyperparameters may only have performed well due to random chance (i.e., the random initialization, action sampling, and environment transitions) during the particular training run, and due to selecting the best result throughout training. Nevertheless, this may not characterize how well different models or hyperparameters may enable learning good policies in general or work in practice. It is therefore important to use separate model hyperparameter selection, training, and evaluation runs for each model. Additionally, it is important to use multiple runs in each case to characterize the distribution of results one can expect with a particular algorithm. We follow such rigorous procedures with our testing framework and in the results reported here.

Specifically, for a given RL algorithm, we first use a common technique of random grid search over the full set of hyperparameters to narrow down a smaller set of key tuning hyperparameters and associated grid of values for a given domain (using multiple random runs with different random seeds per hyperparameter combination). Then, for each environment and RL method, we use multiple random training runs (10 in this paper) for each hyperparameter combination to evaluate each hyperparameter combination. We then select the set of hyperparameters giving the best average result across the multiple random runs as the best hyperparameters to use for that method and environment. Finally, we perform the evaluation runs. For the selected best hyperparameters for a method and environment, we run a new set of 10 random training runs with newly instantiated models to get our final set of trained models. Then, for each of the 10 models, we evaluate the model on multiple randomly initialized episodes of multiple steps (20 episodes and 256 steps used in our experiments) to get a mean reward per model run. Given the set of mean rewards from multiple model runs, we report the overall mean, median, and standard deviation of mean rewards for the particular model (model class and set of best hyperparameters) and environment. This characterizes what kind of performance we would expect to see for that given RL model approach on that environment, and how it varies through random chance due to random initialization and actions/trajectories. This gives a more thorough and accurate evaluation and comparison of the different RL algorithms. Additionally, we performed extensive hyperparameter tuning and focused on bringing out the best performance of the baseline DRL models—searching over 4700 hyperparameter combinations with the initial random grid search to narrow down smaller grids, and using final grids of 32-36 hyperparameter combinations (varying 3 or more hyperparameters) per method to perform final hyperparameter optimization per environment and method. More details on the hyperparameter tuning are provided in \S \ref{subsubsec:hp_tuning}.

\subsection{Performance, policy analysis, parameter tuning and managerial insights}\label{sec:Computations}

\subsubsection{Performance:}
In Table \ref{table:avg_reward}, we present the average per step reward (over test runs) of the different algorithms and compare them to PARL in seven different settings described earlier. We additionally provide the percentage improvement over two widely used methods: PPO and BS. We observe that PARL is a top-performing method; it outperforms all benchmark algorithms in all but one setting, 1\texttt{S}$^{\infty}$-2\texttt{W}-3\texttt{R}. On average, across the different supply chain settings we study, PARL outperforms the best performing RL algorithm by 14.7\% and the BS policy by 45\%.

Notably, the improvements are higher in more complex supply chain settings (1\texttt{S}-20\texttt{R}, 1\texttt{S}-10\texttt{R}, 1\texttt{S}-2\texttt{W}-3\texttt{R}, and 1\texttt{S}-2\texttt{W}-3\texttt{R} (DS)) amongst the settings tested in the paper. While in the 10\texttt{R} and 20\texttt{R} settings, the retailer has to optimize decisions over a larger network with a larger action space, 1\texttt{S}-2\texttt{W}-3\texttt{R} and 1\texttt{S}-2\texttt{W}-3\texttt{R} (DS) are multi-echelon settings with more complex supply chain structures. Similarly, in the 1\texttt{S}-3\texttt{R} setting, the supplier is more constrained than in the 1\texttt{S}-3\texttt{R}-High setting, which makes the inventory allocation decision more complex. PARL's ability to explicitly factor in known state-dependent constraints enables it to outperform other methods in these settings. Settings 1\texttt{S}-3\texttt{R}-High and 1\texttt{S}$^{\infty}$-2\texttt{W}-3\texttt{R} have relatively high supplier production, where BS and related heuristics like DA are known to work well. Here, PARL is on par with the BS heuristic (within one standard deviation of the BS heuristic’s performance) and outperforms other RL methods. In these settings, the combination of surplus inventory availability, coupled with low holding costs and high demand uncertainty, encourages holding high inventory levels over long time horizons. This makes reward attribution for any specific ordering action harder, and thus these settings are harder to learn for RL algorithms.

\begin{table*}[h]
\scalebox{0.665}{\parbox{\textwidth}{
\begin{tabular}{@{}lccccccccc@{}}
\toprule[1.0pt]
%
\textbf{Setting} & \textbf{SAC} & \textbf{TD3} & \textbf{PPO} & \textbf{A2C} & \textbf{BS} & \textbf{DA} & \textbf{PARL} & \textbf{PARL over PPO / BS}  \\\midrule[0.5pt]
1S-3R-High	& \shortstack{	478.8	$\pm$	8.5	\\	478.3	}   	& \shortstack{	374.7	$\pm$	15.7	\\	374.1	}   	& \shortstack{	499.4	$\pm$	5.7	\\	500.2	}   	& \shortstack{	490.8	$\pm$	8.9	\\	490.1	}   	& \bf \shortstack{	513.3	$\pm$	5.9	\\	513.0	}   	&  \shortstack{	474.0	$\pm$	4.6\\ 473.5}
&  \bf \shortstack{	514.8	$\pm$	5.3	\\	514.3	}  & 3.1\% / 0.3\% 	\\\midrule[0.5pt]
1S-3R	& \bf \shortstack{	398.0	$\pm$	3.2	\\	398.3	} 	& \shortstack{	329.6	$\pm$	45.2	\\	311.7	} 	& \bf\shortstack{	397.0	$\pm$	1.6	\\	397.4	} 	& \shortstack{	392.4	$\pm$	4.4	\\	392.87	} 	& \shortstack{	313.7	$\pm$	3.1	\\	314.3	}
&  \shortstack{	303.2	$\pm$	2.2\\ 303.8}
& \bf \shortstack{	400.3	$\pm$	3.3	\\	400.8	} & 0.8 \% / 27.6\%	\\ \midrule[0.5pt]
1S-10R	& \shortstack{	870.5	$\pm$	68.9	\\	905.8	}	& \shortstack{	744.4	$\pm$	71.4	\\	766.3	}	& \shortstack{	918.3	$\pm$	24.7	\\	919.2	}	& \shortstack{	768.1	$\pm$	40.5	\\	773.52	}	& \shortstack{	660.5	$\pm$	2.1	\\	659.9	}
&  \shortstack{	651.9	$\pm$	1.6\\ 652.4}
& \bf \shortstack{	1006.3	$\pm$	29.5	\\	1015.7	} & 9.6\% / 52.3\%	\\ \midrule[0.5pt]
1S-20R	& \bf\shortstack{	1216.1	$\pm$	25.2	\\	1221.3	}	& \shortstack{	1098.0	$\pm$	43.3	\\	1105.8	}	& \shortstack{	1072.6	$\pm$	63.4	\\	1059.3	}	& \shortstack{	1117.3	$\pm$	37.5	\\	1114.4	}	& \shortstack{	862.7	$\pm$	3.0	\\	862.9	}
&  \shortstack{	851.9	$\pm$	1.5\\ 852.2}
& \bf \shortstack{	1379.2	$\pm$	190.1	\\	1434.3	} & 28.5\%  / 59.9\%	\\ \midrule[0.5pt]
1S-2W-3R	& \shortstack{	374.2	$\pm$	3.7	\\	375.0	}	& \shortstack{	361.1	$\pm$	15.4	\\	362.7	}	& \shortstack{	377.5	$\pm$	3.7	\\	377.5	}	& \shortstack{	360.2	$\pm$	23.2	\\	365.3	}	& \shortstack{	300.8	$\pm$	5.4	\\	302.2	}
&  \shortstack{	287.2	$\pm$	1.9\\ 287.1}
& \bf \shortstack{	398.3	$\pm$	2.5	\\	399.7	}& 5.5\%  / 32.4\%	\\\midrule[0.5pt]
1S-2W-3R (DS)	& \shortstack{	344.1	$\pm$	20.6	\\	346.1	}	& \shortstack{	259.3	$\pm$	32.3	\\	262.8	}	& \shortstack{	387.8	$\pm$	5.3	\\	388.9	}	& \shortstack{	327.5	$\pm$	32.7	\\	322.61	}	& \shortstack{	166.2	$\pm$	3.8	\\	166.4	}
&  \shortstack{	158.3	$\pm$	1.7\\ 158.0}
& \bf \shortstack{	405.4	$\pm$	2.0	\\	405.9	} & 4.5\% / 143.9\%	\\
\midrule[0.5pt]
1S$^{\inf}$-2W-3R	& \shortstack{	40.6	$\pm$	59.8	\\	4.2	}	& \shortstack{	21.0	$\pm$	57.0	\\	4.26	}	& \shortstack{	136.8	$\pm$	18.4	\\	136.0	}	& \shortstack{	62.9	$\pm$	33.7	\\	69.1	}	& \bf \shortstack{	208.8	$\pm$	5.2	\\	209.9	}
&  \shortstack{	163.2	$\pm$	4.8\\ 162.6}
& \bf \shortstack{	206.1	$\pm$	9.0	\\	207.9	} & 50.6\% / -1.3\%	\\
 \bottomrule[1.0pt]
 \end{tabular} }}
 \caption{Average per-step-reward with standard deviation and median (next line) of different benchmark algorithms, averaged over different testing runs. We bold all top performing methods: those with performance not statistically significantly worse than the best method, using one standard deviation.}\label{table:avg_reward}
 \end{table*}

We also analyze the rate of learning of different algorithms during training. In Figure \ref{fig:rate_of_learning}, we plot the average per-step reward over training steps from 3 different environments. We find that in each case, the PARL actor performs much worse in the initial, very early training steps on account of optimizing over a poorly trained critic. Once the critic improves in accuracy, PARL is able to recover a very good policy during training.  Furthermore, we generally observe improved sample complexity of PARL compared to the DRL baselines from these learning curves.  That is, we observe that PARL often converges to its final high-reward solution in fewer total steps with the environment (in some cases much fewer steps) compared to the DRL baselines.

\begin{figure*}[h]
\centering
    \begin{subfigure}[b]{0.32\textwidth}
         \centering
         \includegraphics[width=\textwidth]{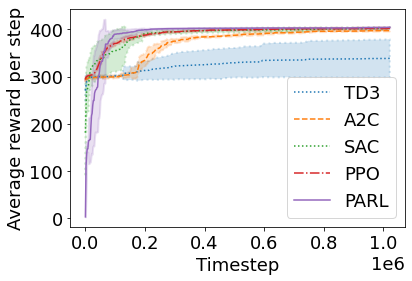}
         \caption{\small{1S-3R}}
         \label{fig:lc_1s3r}
     \end{subfigure}
     \begin{subfigure}[b]{0.32\textwidth}
         \centering
         \includegraphics[width=\textwidth]{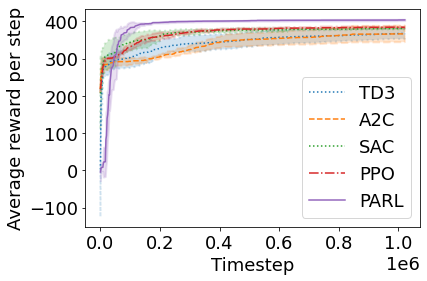}
         \caption{\small{1S-2W-3R}}
         \label{fig:lc_1s_2w_3r}
     \end{subfigure}
     \begin{subfigure}[b]{0.32\textwidth}
         \centering
         \includegraphics[width=\textwidth]{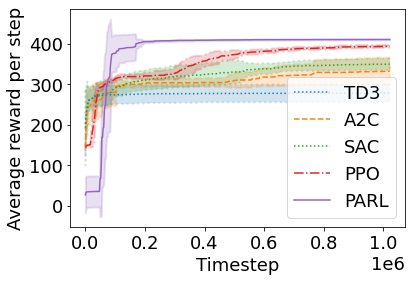}
         \caption{\small{1S-2W-3R (DS)}}
         \label{fig:lc_1s_2w_3r_lo}
     \end{subfigure}
    \caption{Learning curves of PARL and benchmark algorithms during training runs. }
    \label{fig:rate_of_learning}
\end{figure*}
Finally, we report algorithm run-times. The average per-step run time of the PARL algorithm is 0.178, 0.051, 0.050, 0.089, 0.051, 0.044 and 0.042 seconds in the 1\texttt{S}-3\texttt{R}-High, 1\texttt{S}-3\texttt{R}, 1\texttt{S}-10\texttt{R}, 1\texttt{S}-20\texttt{R}, 1\texttt{S}-2\texttt{W}-3\texttt{R}, 1\texttt{S}-2\texttt{W}-3\texttt{R} (DS) and 1\texttt{S}$^{\inf}$-2\texttt{W}-3\texttt{R} settings, respectively. The average per step run time is highest in the 1S-3R-High setting. This is due to the larger feasible action set in this setting. In all other settings, the run time remains below 0.10 seconds. We note that during training, we use 8 parallel environments to gather training trajectories, and use 2 CPLEX threads per environment. The run time can improve further by increasing parallelization. In contrast, the average per-step run time of PPO (the DRL algorithm that performs the best in most settings) is 0.007, 0.005, 0.010, 0.008, 0.008, 0.008, and 0.007 seconds respectively. Clearly, PPO outperforms PARL in terms of run-time. This is because while per-step action in PARL is an outcome of an integer-program, DRL algorithms take gradient steps that are computationally much faster. In Appendix \ref{app:runtime_exps}, we further analyze the scalability of the proposed algorithm as we scale the size of the supply chain or the complexity of the NN. We also test runtime changes if we use open-source IP solvers instead of CPLEX. We find that PARL can be used for moderately sized problems and further discuss other potential directions for scaling the method to solve large-scale problems.

\subsubsection{Analyzing learned replenishment policy from RL algorithms: }
The numerical results from the experiments above show that RL algorithms can provide substantial gains in complex supply chain settings. Naturally, the next important question is how RL algorithms achieve such improved performance and how good these policies are. We answer both these questions in what follows.

\begin{figure*}[h]
\centering
    \begin{subfigure}[b]{0.38\textwidth}
         \centering
         \includegraphics[width=\textwidth]{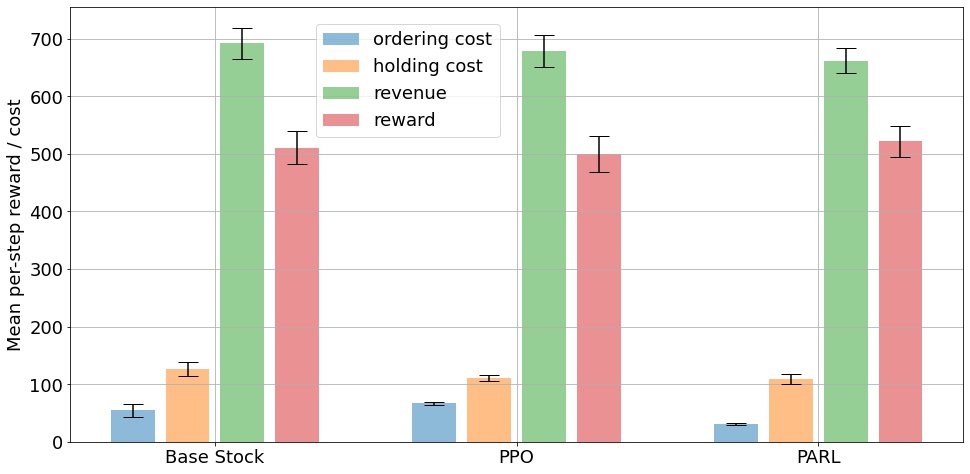}
         \caption{1\texttt{S}-3\texttt{R}-High}
         \label{fig:cost_1s3r-High}
     \end{subfigure}
     \begin{subfigure}[b]{0.38\textwidth}
         \centering
         \includegraphics[width=\textwidth]{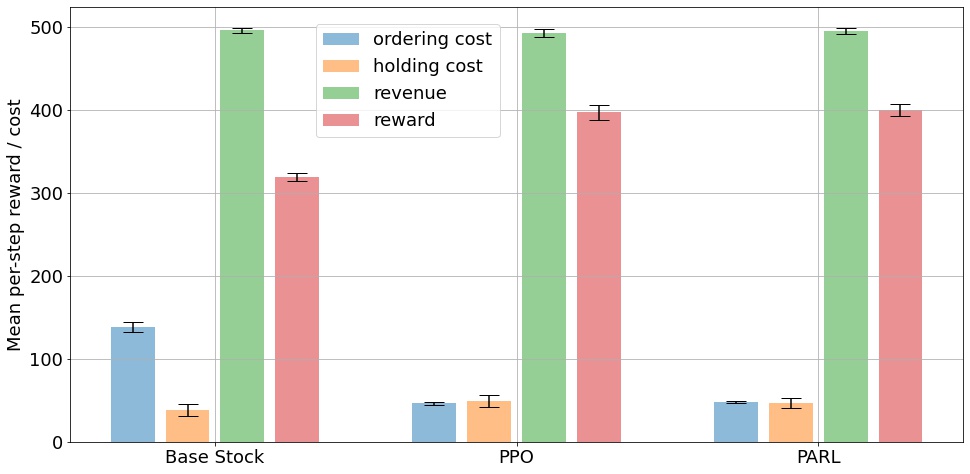}
         \caption{1\texttt{S}-3\texttt{R}}
         \label{fig:cost_1s3r}
     \end{subfigure}
          \begin{subfigure}[b]{0.38\textwidth}
         \centering
         \includegraphics[width=\textwidth]{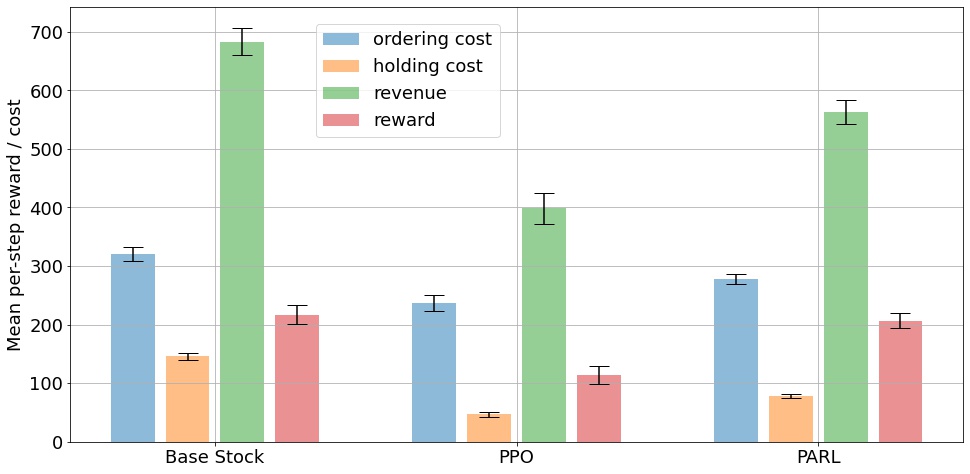}
         \caption{1\texttt{S}$^{\inf}$-2\texttt{W}-3\texttt{R}}
         \label{fig:cost_1sinf2W-3R}
     \end{subfigure}
     \begin{subfigure}[b]{0.38\textwidth}
         \centering
         \includegraphics[width=\textwidth]{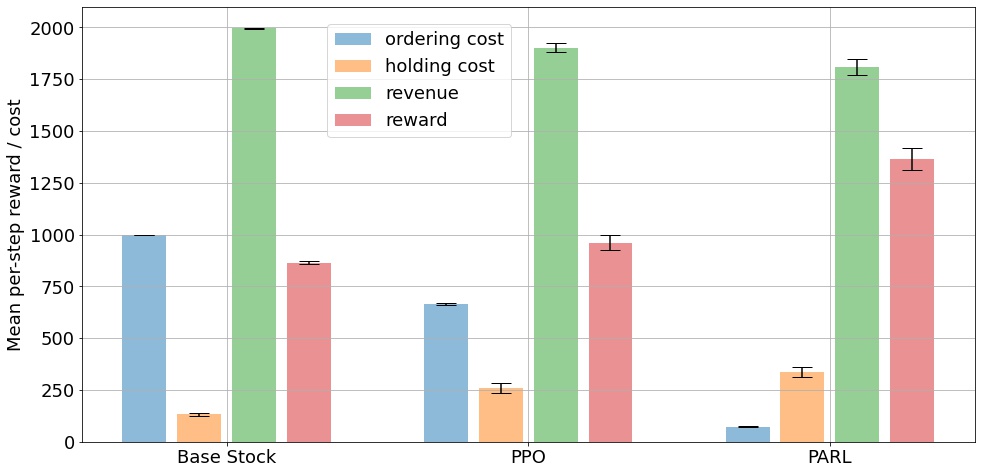}
         \caption{1\texttt{S}-20\texttt{R}}
         \label{fig:cost_1s20r}
     \end{subfigure}
    \caption{Breakdown of rewards in test across BS, PPO and PARL algorithms. Note that ordering costs include fixed and variable costs of ordering, revenue refers to the revenue earned from sales and reward refers to the revenue net costs incurred (see \S \ref{sec:inventory} and Table \ref{table:SCsettings} for more details).   }
    \label{fig:barChartOfCosts}
    \vspace{-2em}
\end{figure*}

\paragraph{Why are RL algorithms performing better in some settings?}
In Figure \ref{fig:barChartOfCosts}, we plot the costs, revenue, and reward breakdown of different algorithms in various supply chain network settings. In the 1\texttt{S}-3\texttt{R}-High setting (see Figure \ref{fig:cost_1s3r-High}), we find that while PARL incurs lower total costs (ordering plus holding), all algorithms perform equally well in terms of overall reward since BS and PPO are able to compensate for higher inventory costs with higher revenue. Similar insights hold in the other high inventory 1\texttt{S}$^{\infty}$-2\texttt{W}-3\texttt{R} setting (see Figure \ref{fig:cost_1sinf2W-3R}), with the exception of the PPO-based policy, whose performance deteriorates compared to the other policies. This is due to lost sales resulting in lower revenue and hence rewards. As we decrease available inventory (see Figure \ref{fig:cost_1s3r}), the performance of the BS policy deteriorates since it incurs high ordering costs due to proportional fulfillment. Note that in this setting, both PARL and PPO are better able to trade off between costs and revenue. As we increase the complexity of the underlying network, PARL's performance improves mainly due to lower costs. In particular, in the largest network with 20 retailers (see Figure \ref{fig:cost_1s20r}), PARL incurs much lower ordering costs compared to the other algorithms. Nevertheless, its holding costs are higher than the other algorithms. This implies that PARL makes fewer orders (but with larger order sizes) which ensure low overall inventory costs. Cost and reward comparisons for other settings are provided in \S \ref{app:benchmark_algos} of the Appendix.







\paragraph{Bench-marking with known optimal policy: }All the settings considered so far in the paper have no analytical optimal solution. Hence, while PARL performs better than known heuristics, its performance might still be far from the optimal policy. Therefore, to compare PARL's performance with an optimal policy, we consider two different settings (one with back-orders and another with lost-sales) where the optimal policy can be computed and compare PARL's performance with the optimal policy.

\noindent \textbf{Back-order setting:} We first consider a simplified setting with one retailer and one supplier with infinite inventory (1\texttt{S}$^{\infty}$-1\texttt{R}) with back-ordered demands, non-zero lead time, and no fixed costs (see Appendix \ref{sec:params1s1r} for full details on the different parameters). In this setting, it is well known that the optimal policy is an order-up-to base-stock policy (i.e., order in each period to maintain an inventory position, which is on-hand plus pipeline, up to this level). The optimal policy is easy to compute as it has a closed-form solution. Therefore, we compare PARL's learned policy with the optimal underlying base-stock policy. For the chosen parameters in this setting, the order-up-to level is 27 (see details of this calculation in \S \ref{sec:params1s1r} of the Appendix). In Figure \ref{fig:1S1R_policy}, we present the learned policy from two different DRL algorithms. On the x-axis, we plot the inventory position (sum of on-hand and pipeline inventory), and on the y-axis, we plot the optimal action taken by the policy. Interestingly, the optimal policy learned by the DRL algorithms is also an order-up-to policy. While this result is not surprising for existing DRL algorithms, especially given the recent success of using off-the-shelf RL methods for inventory management, the analysis highlights the near-optimal performance of PARL and demonstrates that the learned optimal policy mimics the optimal parametric policy in this setting.

\begin{figure*}[h]
\centering
   \includegraphics[width=.45\textwidth]{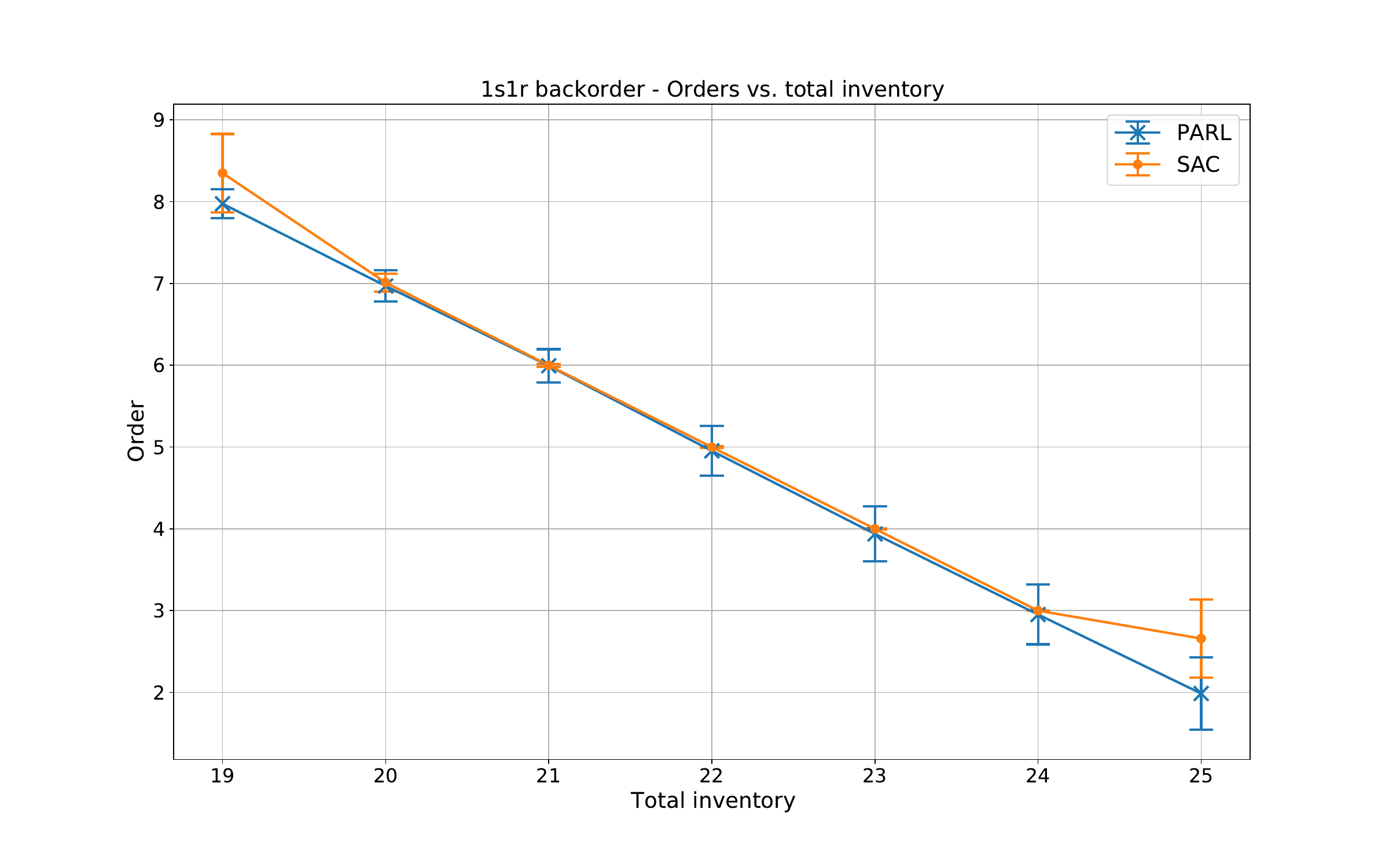}
   \captionsetup{font=small}
         \caption{Comparing the optimal ordering policy of PARL and SAC in the 1\texttt{S}$^{\inf}$-1\texttt{R} backorder setting. Here, the optimal static policy is an order-up to policy with 27 units being the order upto level. Observe that both PARL and SAC are able to learn the optimal order-up-to policy across various on-hand inventory states observed in test (99.9\% of the time). Higher variance in comparison to SAC for PARL can be attributed to its deterministic policy structure that optimizes over the learned critic with a four dimensional input state.   }
         \label{fig:1S1R_policy}
             \vspace{-2em}
\end{figure*}

\begin{figure*}[h]
\centering
   \includegraphics[width=.7\textwidth]{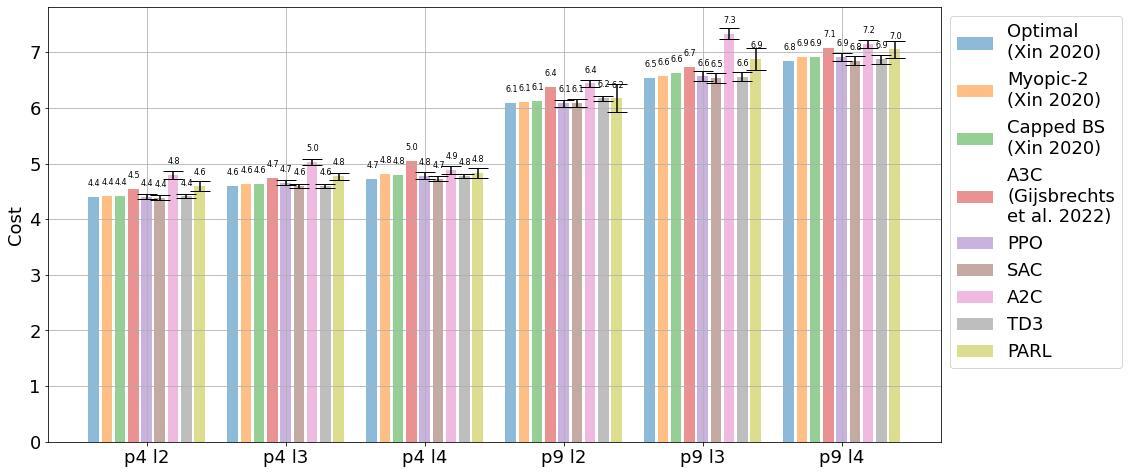}
         \caption{Comparing the performance of various methods in the lost sales setting of \cite{zipkin2008old}. The labels on the x-axis denote different environment settings (e.g., p9 l4 denotes price of 9 and lead time of 4). In this setting, the optimal policy is unknown but near optimal heuristics have been proposed in the past.  PARL, as well as other benchmark DRL algorithms are able to learn \textit{near optimal} policies.   }
         \label{fig:lost_sales_single_retailer}
             \vspace{-2em}
\end{figure*}

\noindent \textbf{Lost-sales setting:}
Finally, we analyze PARL's performance in a lost-sales setting that has been extensively used to benchmark the performance of inventory ordering heuristics in the past (see, e.g., \citealt{zipkin2008old, xin2021understanding, gijsbrechts2018can}). In this setting, we have a single retailer who faces a random demand (Poisson distributed with mean 5), has a unit holding cost, zero ordering cost, and varying selling price. The retailer has to decide how much inventory to order in each period. Inventory is delivered from an upstream node with a lead time. Following \cite{gijsbrechts2018can}, we adopt six experiments where we change the selling price to either 9 or 4, and the lead time of the upstream node to either 2, 3, or 4. For each (selling price, lead time) pair, we run all benchmark RL algorithms, along with PARL, to compare their performance on the average cost metric. Figure \ref{fig:lost_sales_single_retailer} shows the performance of different algorithms in all these settings, including the Capped Base stock and Myopic 2 policies of \cite{xin2021understanding} and the DRL-based A3C policy of \cite{gijsbrechts2018can}. We find that PARL, as well as benchmark DRL methods, are able to learn near-optimal inventory policies. PARL's optimality gap is between 1\%-5\% in various settings, while it is within 1\% for PPO and SAC (note that PARL and benchmark algorithms are tested with 10K steps in the testing phase). Our findings are similar to those reported in previous studies: in settings where near-optimal heuristics exist, DRL-based methods are able to learn \textit{very good} policies, but they are not able to outperform the best-performing heuristics. These results provide an expanded set of benchmarks for this setting.

Finally, in Appendix \ref{app:acyclic}, we present additional experiments on serial acyclic chains and find similar insights: the performance of PARL (as well as other RL methods) is comparable (within 0\% - 5\%) of the optimal Clark and Scarf based policy \citep{clark1960optimal}.





 \subsubsection{Hyper-parameter tuning of RL algorithms: }\label{subsubsec:hp_tuning} Given the number of hyperparameters that need to be selected for running RL algorithms, we divide them into two sets: (i) parameters that were fixed to be the same across different benchmark algorithms and (ii) parameters that were tuned as we trained different benchmark algorithms.

\paragraph{Fixed hyperparameters:}
Here we report the fixed set of hyperparameters used by all methods. We fix batch size, the NN architecture, the neuron activation function, and the state space representation\footnote{\label{footnote:reps}States and actions can be represented in different ways, which affect modeling outcomes and so can be treated as a hyperparameter. Discrete means each possible state (i.e., combination of inventory values) or action (i.e., combination of order actions per entity) is given a unique index (so a policy network would output a probability for each possible index - i.e., state or action variable value combination) and a policy network gives a probability for each discrete action. Multi-discrete means each different state or action variable is encoded with its own set of discrete values and the policy network gives a probability for each state of each variable (e.g., output a probability over a discrete set for each order action to be taken). Finally, continuous (and normalized continuous) means the state and action variables are treated as continuous values - so the policy network outputs a single value for each action variable (i.e., each order action) and the probability is modeled via a continuous distribution, most often Gaussian.}, the action representation\footnotemark[\getrefnumber{footnote:reps}], as well as the epoch length for generating trajectories. These were determined based on two factors: (1) the commonly used settings across the RL literature (for example, 64x64\footnote{64x64 represents a NN with two layers, each of fully connected 64 neurons.} NN architecture and batch size of 64 is most commonly used across many different problems and methods), and by sampling random combinations from a large grid of hyperparameters and comparing results trends to narrow down the set of hyperparameters to consider to consistently well-performing values and reasonable ranges.

As such, this was an iterative process where we tried a range of hyperparameters, then refined. For example, for the network architecture size, we also tried larger sizes including 128x128, 512x512, 1024x1024, 128x128x128, 512x256, 1024x512, 1024x512x256, 128x32, 512x128, and 512x256x64. Nevertheless, larger sizes did not see considerable improvement but led to a significant increase in runtime. Similarly, we tried batch sizes of 32, 64, and 128, but found no significant differences in performance and hence selected the most commonly used 64 as the batch size. To select the activation function, we tested both ReLU and tanh activation. We found that ReLU activation performs as well or better than tanh (overall it gave close but slightly better results). Hence, we fixed the activation to ReLU across methods for fair comparison, and for ease of implementation with more efficient optimization methods \citep{glorot2011deep, hara2015}. Similarly, for the state and action space representation, we tested discrete, multi-discrete, and continuous representations (normalized between -1 and 1). We found that both continuous state and action representations work best in these settings.

Finally, we also comment on some algorithm-specific hyperparameters. For the number of internal training iterations per collected buffer (PPO-specific setting), we tried 10, 20, 40, and 70 and found the best results with 10-20, so fixed this parameter to 20. For the number of steps per epoch/update (PPO and A2C-specific setting), we tried 512, 1024, and 2048 and found the larger number to give better results generally, so fixed this to 2048. For PPO, we also found that early stopping the policy update per epoch, based on a KL-divergence threshold of default 0.15, consistently provided better results than not using this. The final set of fixed hyperparameters used across all experiments, models, runs, and environments are given in Table \ref{tab:fixed_hps} of Appendix \S \ref{appendix:parameter_tuning}.

\paragraph{Hyper-parameter search:} Next, we discuss the set of hyper-parameters that were tuned for each supply chain setting during training for different algorithms. Table \ref{tab:tune_hps_ppo_a2c} of Appendix \S\ref{appendix:parameter_tuning} describes the set of parameters for PPO and A2C. Similarly, Table \ref{tab:tune_hps_sac_td3} of Appendix \S\ref{appendix:parameter_tuning} describes the set of parameters for SAC and TD3. For PARL, a common set of hyper-parameters was used across all settings, with the exception of the discount factor, $\gamma$, for the infinite supplier inventory setting, since we observed universally higher gamma being necessary for the baseline DRL methods (see Table \ref{tab:best_hps} of Appendix \S\ref{appendix:parameter_tuning}). The discount factor was set to 0.75 (0.99 for the infinite supplier inventory setting only), the learning rate was set to 0.001, and the sample-averaging approach used was quantile sampling with 3 demand samples per step. In each case, this set of hyper-parameters was then used for the evaluation of the RL model by retraining 10 different times with different random seeds using those best hyper-parameters for each method, and reporting statistics on the 20-episode evaluations of the best epoch model across the 10 runs. Besides the parameters mentioned here, all other parameters were set at their default values in the Stable Baselines 3 implementation (see API\footnote{\url{https://stable-baselines3.readthedocs.io/en/master/}} for more details). Note that our experiments revealed that the standard gamma value of $0.99$ consistently gave much poorer results than smaller gamma values for most environments (we experimented with $0.99$, $0.9$, $0.85$, $0.8$, and $0.75$) - so we included smaller gamma values in our hyper-parameter grids but still kept the option of the traditionally used $0.99$ gamma for the benchmark RL methods in case they were able to factor in longer-term impact better. Finally, note that default optimizers are used: ADAM \citep{kingma2015adam} for all except A2C which uses RMSprop \citep{hinton2012rmsprop} by default.

\paragraph{Evaluation of hyperparameters:}

In Table \ref{tab:best_hps} of Appendix \S \ref{appendix:parameter_tuning}, we show the selected set of best hyperparameters used for each benchmark RL method and environment. These were selected based on the parameter settings that gave the best average reward (maximum over the training epochs), averaged across 10 different model runs for that hyperparameter combination.

\subsubsection{Managerial Insights} \label{subsec:managerial_insights}
We have so far extensively tested the proposed PARL methodology to compare its performance with state-of-the-art benchmark methods. In what follows, we present the key insights from these experiments that could be of interest to OR practitioners.
(i) The proposed PARL framework is most appropriate in settings with supply chain networks that neither have easy-to-compute optimal policies nor good heuristic solutions. PARL is better able to leverage the inherent structure of inventory management problems (by modeling the immediate objective value and state transitions based on inventory dynamics) and combine it with the predictive power of deep RL (by using NN-based value estimators). Importantly, it also outperforms out-of-the-box RL heuristics by leveraging math-programming techniques to optimize actions. This is evident in our numerical experiments, particularly in the settings discussed in Table 2 with more complex supply chain structures and limited inventory availability. (ii) In other settings where either optimal policies or near-optimal heuristics are available (for example, order-up-to heuristics and the back-order and lost-sales settings discussed in the preceding section), PARL's (and other RL methods) overall performance is comparable to benchmark methods. Hence, one could directly leverage these heuristics for inventory optimization on account of reduced computational complexity in comparison to PARL and other RL methods. (iii) As is the case with RL-based techniques, the overall performance of PARL relies on hyper-parameter tuning and testing and requires significant computational resources.
To further test the scale of problems that PARL can tackle, in Appendix \ref{app:runtime_exps}, we perform extensive numerical experiments to explore how the run-time changes as we (a) scale the supply chain network; (b) increase the size of the NN used for value approximation; or (c) use open-source IP solvers instead of CPLEX.  Our computational experiments show that PARL can be trained to solve moderate to large sized supply chain problems (involving hundreds of retailers) in approximately 10 days of training time given the computing resources we used per run. While this is practical in scenarios with stationary dynamics, additional effort is required to make the framework scalable and practical in other settings. We discuss potential directions for further speed ups in \S5. To further simplify this overall training process and make it more accessible to practitioners, we discuss the details of an open-source Python library.

\subsection{Open source tool for bench-marking}\label{subsec:bench_marking}

In this section, we briefly discuss the development and usage of an open-source Python library that can facilitate the development and benchmarking of reinforcement learning on diverse supply chain problems\footnote{See \url{https://github.com/divyasinghvi/PARL-InventoryManagement.git.} for more details}. The library is designed to make it easy to define arbitrary, customizable supply chain environments (such as those used in this paper) and to easily plug in different RL algorithms to test them on the environments. Different supply chain configurations can easily be defined via code or configuration files, such as varying supply chain network structures, capacities, holding costs, lead times, demand distributions, etc. For example, the configuration file for the library defining the 1\texttt{S}-2\texttt{W}-3\texttt{R} supply chain environment used in this paper, with one supplier, two warehouses, and three retailers, under a lost sales setting is shown in Listing \ref{listing:pdr_1s2w3r_ds_cfg} of Appendix \S\ref{app:opensource_listings}.

This library is based around the OpenAI Gym environment API \citep{brockman2016}, so that standard RL implementations, like Stable Baselines 3 \citep{stable-baselines3}, can work directly with defined environments. Listing \ref{listing:ex_dnv_usage} in Appendix \S\ref{app:opensource_listings} shows an example of instantiating the 1\texttt{S}-2\texttt{W}-3\texttt{R} environment and exercising it—taking actions and getting rewards and next states from the environment, and loading and training a DRL baseline with the environment. This also demonstrates how the library supports easily specifying different representation encodings for the states and actions (common representation as mentioned in an earlier footnote: discrete, multi-discrete, continuous, and normalized continuous). Additionally, this example illustrates enabling logging so all states and inventory levels, actions and orders, rewards, demands, and costs per network entity are logged and can be exported after taking actions in the environment.

To the best of our knowledge, such a reinforcement learning library and set of environments focused on broadly enabling supply chain environments for reinforcement learning does not currently exist. OpenAI provides the Gym library which provides an interface for RL environments and a set of video game environments. Various libraries exist with RL algorithms that leverage video game environments. However, there are no libraries focused on RL for supply chain management that facilitate easy creation of different supply chain management environments. There is an open-source library that provides a set of general environments for a variety of operations research problems, ORGym, but it is more broadly focused on OR as a whole and only includes a couple of environments for supply chains under specific settings. We believe our framework is complementary to this and essentially provides a full set of environments for common supply chain scenarios as well as the capability to flexibly define custom environments and specify environment variations of interest, from different network, cost, and fulfillment structures, to different demand distributions including non-stationary and supply chain settings such as the lost sales setting.

\section{Conclusions and future research directions}

Reinforcement learning has led to considerable breakthroughs in diverse areas such as robotics, games, and many others in the past decade. In this work, we present an RL-based approach to solve some analytically intractable problems in supply chain and inventory management. Many real-world problems have large combinatorial actions and state-dependent constraints. Hence, we propose the PARL algorithm that uses integer programming and SAA to account for underlying stochasticity and provide a principled way of optimizing over large action spaces. We then discuss the application of PARL to inventory replenishment and distribution decisions across a supply chain network. We demonstrate PARL's superior performance on different settings which incorporate some real-world complexities including heterogeneous demand across the demand nodes in the network, supply lead times, and lost sales. Finally, to make the work more accessible, we also detail the development of a Python library that allows easy implementation of various benchmark RL algorithms along with PARL for different inventory management problems.

This work also opens up various avenues for future research. First and foremost, by making RL algorithms more accessible via the Python library, we believe that researchers will be able to easily benchmark and also propose new RL-inspired algorithms for other complex supply chain problems. For example, one potential interesting research direction is to use this methodology in settings where the complete demand distribution is unknown. In these cases, one can use historical sales data to estimate demand distributions and improve demand forecasts over time using online learning techniques. The improved forecasts can then be directly used for generating replenishment policies by leveraging the proposed framework. Another interesting direction could be to extend this framework for order fulfillment, wherein the demand is realized and the decision is where to fulfill it from. Similarly extending the IP based formulation to other NN architectures with different activation functions could be an interesting avenue for future research, Finally, since IP-based methods still suffer from longer run times and use significant computational resources, coming up with alternate near-optimal formulations that are computationally efficient could be very useful. For example, further speedups could be possible via approximations or decomposition or designing new NN architectures specifically for inventory optimization, an interesting direction for future research.

 \bibliographystyle{informs2014}
{\SingleSpacedXI
\bibliography{BIB-ITE.bib}
}
\newpage
\begin{APPENDICES}
\section{Table of notation and block diagram for PARL}\label{app:notation_and_block}
\begin{figure}[h]
\centering
\includegraphics[width=.6\textwidth]{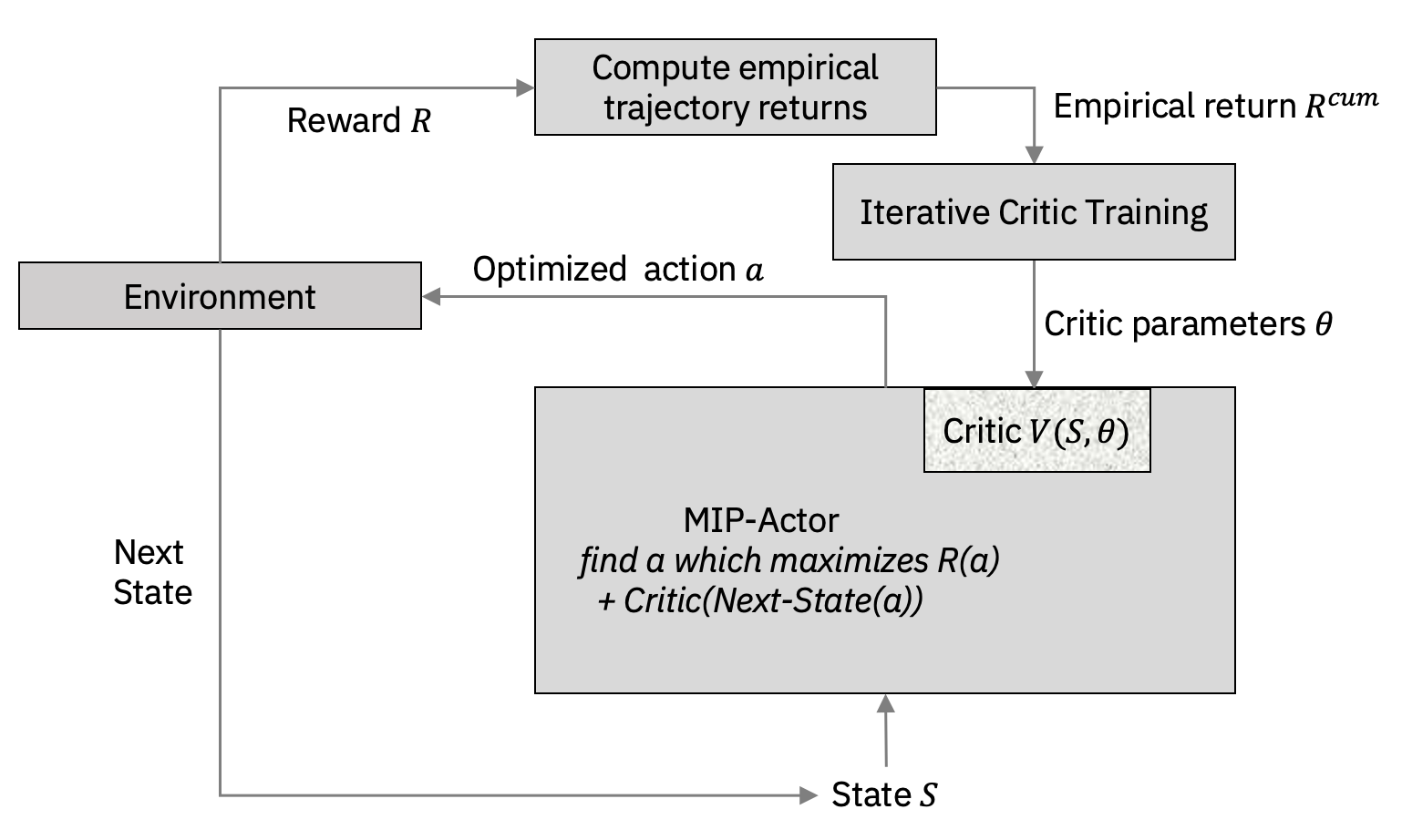}
\caption{Block diagram illustration of the PARL algorithm}
\label{fig:PARL}
\end{figure}
\begin{table}[H]
\centering
\scalebox{0.78}{\parbox{\textwidth}{
\begin{tabular}{|c|l|}
\hline
\textbf{Symbol} & \textbf{Explanation} \\
\hline
$s$ & State in the state space $\mathcal{S}$. \\
\hline
$a$ & Action in the action space $\mathcal{A}(s)$, associated with state $s$. \\
\hline
$D$ & Uncertain random variable representing demand, taking values in $\mathbb{R}^{\texttt{dim}}$. \\
\hline
$P(D=d\mid s)$ & Probability distribution of $D$ given the context state $s$. \\
\hline
$R(s, a, D)$ & Reward function, dependent on the state $s$, action $a$, and demand $D$. \\
\hline
$\beta$ & Distribution over initial states. \\
\hline
$\gamma$ & Discount factor for future rewards. \\
\hline
$s'$ & Next state, resulting from the transition dynamics $\mathcal{T}(s,a,d)$. \\
\hline
$\pi$ & A stationary policy, a distribution over actions given state $s$. \\
\hline
$J^{\pi}$ & Expected return of a policy $\pi$, based on initial state distribution $\beta$ and value function $V^{\pi}(s)$. \\
\hline
$V^{\pi}(s)$ & Value function under policy $\pi$, sum of discounted rewards starting from state $s$. \\
\hline
$\pi^*$ & Optimal policy that maximizes the long-term expected discounted reward. \\
\hline
$\mathbb{E}_D$ & Expectation taken over the distribution of demand $D$. \\
\hline
$\theta$ & Parameters of the neural network representing the value-to-go function. \\
\hline
$R^{cum,n}_t$ & Cumulative discounted reward from time $t$ in sample path $n$, under policy $\pi_{j-1}$. \\
\hline
$\eta$ & Number of independent realizations of the uncertainty $D$ used in the SAA approach. \\
\hline
$\hat{V}^{\pi}_{\theta}(s)$ & Approximation of the value function at state $s$ under policy $\pi$, parameterized by $\theta$. \\
\hline
$\hat{\pi}^{\eta}_{j}(s)$ & Policy obtained by solving the SAA approximation problem at iteration $j$ for state $s$. \\
\hline
$\Lambda$ & Set of nodes in the supply chain network. \\
\hline
$l$ & Index for nodes in the supply chain network (warehouses, distribution centers, retail stores). \\
\hline
$D^{p}_l$ & Random variable denoting inventory produced at node $l$. \\
\hline
$D^{d}_l$ & Random variable denoting demand generated at node $l$. \\
\hline
$O_l$ & Set of upstream nodes that can ship to node $l$. \\
\hline
$L_{ll'}$ & Deterministic lead time for trans-shipment from node $l$ to node $l'$. \\
\hline
$K_{ll'}$ & Fixed cost associated with trans-shipment from node $l$ to node $l'$. \\
\hline
$C_{ll'}$ & Variable cost associated with trans-shipment from node $l$ to node $l'$. \\
\hline
$h_l$ & Holding cost per inventory unit at node $l$. \\
\hline
$p_l$ & Profit (or revenue) per inventory unit sold at node $l$. \\
\hline
$I^0_l$ & On-hand inventory at node $l$. \\
\hline
$\mathtt{tsc}_l$ & Trans-shipment cost for node $l$. \\
\hline
$\tilde{I}_l^0$ & Intermediate on-hand inventory at node $l$. \\
\hline
$\mathtt{rs}_l$ & Revenue from sales for node $l$. \\
\hline
$\mathtt{hsc}_l$ & Holding-and-salvage costs for node $l$. \\
\hline
$\bar{U}_l$ & Storage capacity of node $l$. \\
\hline
$x_{l^{\prime}l}$ & Trans-shipment decision, inventory shipped from node $l^{\prime}$ to $l$. \\
\hline
\end{tabular}}}
\caption{Mathematical Notations and Their Explanations}
\label{table:notations}
\end{table}

\section{Proof of Proposition \ref{prop:SAA_guarantee} }\label{app:SAA_guarantee}

Recall that our objective is to prove the following result:

\textbf{Proposition 1:} Consider epoch $j$ of the PARL algorithm with a ReLU-network value function estimate $\hat{V}_{\theta}^{\pi_{j-1}}(s)$ for some fixed policy $\pi_{j-1}$. Suppose $\pi_j, \hat{\pi}^{\eta}_j$ are the optimal policies as described in Problem (\ref{eq:per_step_problem2}) and its corresponding SAA approximation respectively. Then $\forall~ s$,
\[\lim_{\eta\to\infty}\hat{\pi}^{\eta}_j(s) = \pi_j(s).\]

To prove this result, we introduce some more notation. Consider the following problem:
\[\min_{x\in\mathcal{X}} \Big\{f(x):=\mathbb{E}[F(x,\varepsilon)]\Big\}\,,\]
where $\mathcal{X}$ is non-empty closed subset of $R^d$, $\varepsilon$ is a random variable with finite-support. Also let $\varepsilon_1,..,\varepsilon_N$ denote $N$ independent realizations of the random variable $\varepsilon$ and define the Sample Average Approximation of the true problem as
\[\min_{x \in \mathcal{X}}\Big\{\hat{f}_N(x) := \frac{1}{N}\sum_{i=1}^NF(x,\varepsilon_i)\Big\}\,.\]
Also let $v^*$ and $S^*$ to be the optimal value and optimal solution of the true problem and $\hat{V}_N$ and $\hat{S}_N$ to be the optimal solution for the SAA problem. Finally, let $\mathcal{D}(S_1,S_2)$ denote the discrepancy between two sets, $S_1$ and $S_2$.

Our main result uses the following two key results from the SAA literature:
\begin{theorem}[Theorem 5.3 of \cite{shapiro2014lectures}]\label{thm:shapiro2}
Suppose that there exists a compact set $\mathcal{C}\subset \mathbb{R}^n$ such that (i) the set of optimal solutions of the true problem is non-empty and is contained in $\mathcal{C}$, (ii) the function $f(x)$ is finite valued and continuous on $\mathcal{C}$, (iii) $\hat{f}_N(x)$ converges to $f(x)$ w.p. 1 as $N \to \infty$, uniformly in $x \in \mathcal{C}$, (iv) for N large enough the set $\hat{S}_N$ is non-empty and $\hat{S}_N \subset  \mathcal{C}$. Then, $\hat{v}_N \to v^*$ and $\mathcal{D}(\hat{S}_N, S^*) \to 0$ w.p. 1 as $N \to \infty$.
\end{theorem}

\begin{proposition}[Proposition 7 of \cite{shapiro2003monte}]\label{prop:shapiro2}
Let $\mathcal{C}$ be a nonempty compact subset of $\mathbb{R}^n $and suppose that: (i) for almost every $\varepsilon$, $F(x,\varepsilon)$ is continuous on $\mathcal{C}$, (ii) $F(x,\varepsilon)$, $x \in \mathcal{C}$, is dominated by an integrable function, (iii) the sample is iid. Then the
expected value function f(x) is finite valued and continuous on $\mathcal{C}$, and $\hat{f}_N(x)$ converges to $f(x)$ w.p. 1 as $N \to \infty$, uniformly in $x \in \mathcal{C}$.
\end{proposition}
\textit{Proof of Proposition \ref{prop:SAA_guarantee}}: Recall that
\begin{equation}\label{eq:per_step_problem2}
\pi_{j}(s) = \arg\max_{a \in \mathcal{A}(s)} \mathbb{E}_D\left[R(s,a,D)+ \gamma \hat{V}^{\pi_{j-1}}_{\theta}(\mathcal{T}(s,a,D))\right].
\end{equation}
and its corresponding SAA version is given by
\begin{equation}\label{eq:SAA2}
\hat{\pi}^{\eta}_{j}(s) = \arg \max_{a \in \mathcal{A}(s)}\frac{1}{\eta}\sum_{i=1}^\eta R(s,a,d_i)+ \gamma \hat{V}^{\pi_{j-1}^{\eta}}_{\theta}(\mathcal{T}(s,a,d_i))\,.
\end{equation}

Let $\hat{f}^s_{\eta}(a):= \frac{1}{\eta}\sum_{i=1}^{\eta} g(s,a,d_i), ~\forall s $, where $g(s,a,d) = R(s,a,d)+ \gamma \hat{V}^{\pi_{j-1}}_{\theta}(\mathcal{T}(s,a,d))$. we can then leverage results of Theorem \ref{thm:shapiro2} as long as $g(s,a,d)$ satisfies the conditions stated in Theorem \ref{thm:shapiro2}.

Hence, we need to prove (i) the set of optimal solutions of (\ref{eq:per_step_problem2}) is non-empty and is contained in $\mathcal{C}$, (ii) the function $\mathbb{E}[g(s,a,D)]$ is finite valued and continuous on $\mathcal{C}$, (iii) $\hat{f}^s_\eta(a)$ converges to $\mathbb{E}[g(s,a,D)]$ w.p. 1 as $\eta \to \infty$, uniformly for $a \in \mathcal{C}$, (iv) for $\eta$ large enough, the set $\hat{S}_{\eta}$ is non-empty and $\hat{S}_{\eta} \subset  \mathcal{C}$.

We start by focusing on property (iii) and show that $\hat{f}^s_\eta(a)$ uniformly converges to $\mathbb{E}[g(s,a,D)]$ with probability 1. Consequently, to prove this result, we prove two main properties of $g(s,a,d)$: (\texttt{p}-i) $g(s,a,d)$ is \textit{continuous} in $a$ for almost every $d \in D$, and (\texttt{p}-ii) $g(s,a,d)$ is dominated by an \textit{integrable function}. To prove (\texttt{p}-ii), we show that $g(s,a,d) \leq C<\infty$ w.p. 1 $\forall a \in \mathcal{A}(s)$. First, notice that $g(s,a,d)$ is an affine function of the immediate reward $R(s,a,d)$ and NN approximation of the value-to-go function. By assumption, the immediate reward follows properties (\texttt{p}-i) and (\texttt{p}-ii). Hence, to show these properties for $g(s,a,d)$, we only need to illustrate that the value-to-go estimation also follows these properties. Consider the value-to-go approximation, simply denoted as $\hat{V}_{\theta}(\mathcal{T}(s,a,d))$ with $\theta= (c,\{(W_k, b_k)\}_{k=1}^{K-1})$ denoting the parameters of the $K$-layer ReLU-network. As $\mathcal{T}(s,a,d)$ is continuous and $\hat{V}_{\theta}(s)$ is continuous, $\hat{V}_{\theta}(\mathcal{T}(s,a,d))$ is continuous which proves (\texttt{p}-i). Next, note that $\mathcal{T}(s,a,d)$ lies in a bounded space for any realization of the uncertainty $d$. Furthermore, since the parameters of the NN ($\theta$) are bounded, the outcome of each hidden layer, and subsequently the outcome of the NN are also bounded. This proves that the NN is uniformly dominated by an integrable function which proves (\texttt{p}-ii). Then, following Proposition \ref{prop:shapiro2}, we have uniform convergence of $g^\eta(s,a,d)$ to  $\mathbb{E}[g(s,a,D)]$ w.p. 1. This proves property (iii). Properties (i), (ii) and (iv) are a direct consequence of the assumptions we make on the solution of Problem (\ref{eq:SAA2}). In particular, for all $s$ the set of feasible actions is a bounded polyhedron $\mathcal{A}(s)$ and that for any $\eta$, the set of optimal actions $\hat{\pi}^{\eta}(s)$ is non-empty. This proves the final result.
$\square $

While the above proof assumes that the action space is continuous, one can extend the results in the case of discrete action spaces as well. See \cite{kim2015guide} for a discussion on the techniques used for extending the analysis to this setting.

\clearpage
\section{Simplified example to demonstrate PARL's application in the inventory replenishment context}\label{app:simplified_inv_example}

In this section, we discuss a simplified supply chain example to demonstrate how the PARL framework is applied to solve the inventory replenishment problem.

\textit{Supply chain: }We consider a single-echelon supply chain with lost-sales where a single producer serves two retailers (see Figure \ref{fig:NN_small_SC_Example}). The two retailers have identical demand but are heterogeneous with respect to the lead time as well as the holding cost. While the first retailer has a lead time of 1 and holding cost of \$1/unit/time, the second retailer has a lead time of 2 and holding cost of \$2/unit/time. Both retailers have a holding capacity of 50 units, a fixed order cost of \$50/order and finally earn a revenue of \$50/unit. Finally, the producer has no holding cost, produces 70 units/time and has a holding capacity of 70 units.

The state-space of this supply chain is the on-hand and pipeline inventory of each retailer and the supplier. It is a six-dimensional vector \textbf{I} = [$s_1, s_2,s_3,s_4,s_5$,$s_6$] where $s_1$ represents the on-hand inventory of the produces, the next two positions contain the on-hand and pipeline inventory of retailer one and the last three positions contain the on-hand and pipeline inventory of retailer two. Now consider any time $t$ such that the initial state is $s$, the order decision is $(x_1, x_2)$ and the demand realization is $(d_1, d_2)$. Our approach considers splitting the total reward into two parts: immediate reward (from this time step) and the value-to-go of future rewards (generated from state transition due to current action and the subsequent actions) which is estimated using a neural network. Consider a simplified setting where the future reward is estimated using a neural network with an input layer, a single hidden layer with two neurons each, and the output layer (See Figure \ref{fig:NN_small_SC_NN_Example2}). The current state \textbf{I}, action $a$ and demand realization $d$ leads to state transition from \textbf{I} to $\textbf{I}^{\prime}$ whose value can be estimated using the NN. In particular, $\textbf{I}^{\prime}$ is a six-dimensional input vector that is passed first through the neurons of the input layer, then to the intermediate layer, and finally transformed to the output which is the estimated value-to-go.
To describe the corresponding IP formulation, consider the first neuron of the input layer. The NN is pre-trained. Hence each neuron of the input layer has corresponding weights (6 dimensional weights $w$) and bias (scalar $b$) and uses the ReLU activation. We denote by ($w_{11}, b_{11}$) and ($w_{12}, b_{12}$), the weight and bias parameters of these neurons in the input layer. Then,
passing the input $\textbf{I}^{\prime} \in [l,u]$ to these neurons, the output from these neurons is given by $z^*_{11}$ and $z^*_{12}$ where
\begin{equation*}
\begin{aligned}
 &(z_{11}^*,y_{11}^*)= \text{arg max}_{z,y}~~ 0\\
 &z \geq w_{11}^T \textbf{I}^{\prime} + b_{11}, ~
z \geq 0, ~
z \leq w_{11}^T \textbf{I}^{\prime} + b_{11} - M^-(1-y),~
z \leq M^+y,
~y \in \{0,1\}\,.
\end{aligned}\label{eq:NN_representation_11}
\end{equation*}
Here, $[l,u]$ are upper and lower bounds on the state and $M^+$ and $M^-$ are big $M$s that can be pre-calculated (see \S 2.2 of the manuscript). To see the intuition about this formulation, note that for any scalar input $u\in \mathbb{R}$, the ReLU activation function is given by $g(u) := \text{max} \{0,u\}$. Hence, if  $w_{11}^T \textbf{I}^{\prime} + b_{11}$ is positive, we want $z_{11}^*$ to equal $w_{11}^T \textbf{I}^{\prime} + b_{11}$ and 0 otherwise. In fact, the output of the system of equations is always a singleton. It is easy to check that the output of the IP above ensures these outputs. Similarly, for the second neuron in the input layer, we have that
\begin{equation*}
\begin{aligned}
 &(z_{12}^*,y_{12}^*)= \text{arg max}_{z,y}~~ 0\\
 &z \geq w_{12}^T \textbf{I}^{\prime} + b_{12}, ~
z \geq 0, ~
z \leq w_{12}^T \textbf{I}^{\prime} + b_{12} - M^-(1-y),~
z \leq M^+y,
 ~y &\in \{0,1\}\,.
\end{aligned}\label{eq:NN_representation_12}
\end{equation*}
The bounded output from the input layer ($z_{11}^*, z_{12}^*$) becomes an input to the neurons of the second layer (denoted by $\text{NN}_{21}$ and $\text{NN}_{22}$ in the figure). We can use similar IP formulation to pass the output from the previous layer (input layer in this case) to this layer. For example, for $\text{NN}_{21}$, we have that
\begin{equation*}
\begin{aligned}
 &(z_{21}^*,y_{21}^*)= \text{arg max}_{z,y}~~ 0\\
 &z \geq w_{21}^T [z_{11}^*, z_{12}^*] + b_{21}, ~
z \geq 0, ~
z \leq w_{21}^T [z_{11}^*, z_{12}^*] + b_{21} - M^-(1-y),~
z \leq M^+y,
 ~y \in \{0,1\}\,.
\end{aligned}\label{eq:NN_representation_21}
\end{equation*}
Similarly, for $\text{NN}_{22}$, we have that
\begin{equation*}
\begin{aligned}
 &(z_{22}^*,y_{22}^*)= \text{arg max}_{z,y}~~ 0\\
 &z \geq w_{22}^T [z_{11}^*, z_{12}^*] + b_{22}, ~
z \geq 0, ~
z \leq w_{22}^T [z_{11}^*, z_{12}^*] + b_{22} - M^-(1-y),~
z \leq M^+y,
 ~y \in \{0,1\}\,.
\end{aligned}\label{eq:NN_representation_22}
\end{equation*}
We let \texttt{P}($w$,$b$,$x$) denote the constraints of the IP for a neuron with weights $w$, bias $b$ and input $x$ to concisely represent the constraint sets in each of these problems. Finally, the output from the hidden layer is passed on to the output layer for final value-to-go estimation of the input state $s'$:
\begin{equation}\label{eq:value_to_go_AE}
\begin{aligned}
V(\textbf{I}^{\prime}) &= c^T [z_{21}^*,z_{22}^*] + b_{o}\\
s.t. ~ &(z_{1j}^*,y_{1j}^*) \in \texttt{P}(w_{1j},b_{1j},\textbf{I}^{\prime}) ~\forall j \in [1,2]\\
~ & (z_{2j}^*,y_{2j}^*) \in \texttt{P}(w_{2j},b_{2j},[z_{11}^*,z_{12}^*]) ~\forall j \in [1,2]\\
\end{aligned}
\end{equation}
where $c \in \mathbb{R}^2$ and $b_o \in \mathbb{R}$ are pre-trained parameters corresponding to the output layer.

Note that the input $\textbf{I}^{\prime}$ is directly a function of the current action (along with the current state $s$ and the demand realization $d$). Let $\textbf{I}^{\prime}: = \mathcal{T}(\textbf{I},\textbf{x},\textbf{d})$ denote the next state given the current state $\textbf{I}$, the action $\textbf{x}$ and the demand realization $\textbf{d}$. Then, the optimal action maximizes both the immediate reward as well as the future cumulative discounted rewards from this action (estimated using the NN described above). In particular, given the current state $\textbf{I}$, we solve the following optimization problem per time step
\begin{align}\label{eq:per_step_SAA_AE}
\textbf{x}^* = \max_{\textbf{x} \in \mathcal{L}(\textbf{I})}
                  & \frac{1}{2} \sum_{i=1}^2 \left[ \overset{*}R(\textbf{I}, \textbf{x},\textbf{d}_i) + \gamma V(\textbf{I}^{\prime}) \right]\,,
\end{align}

where we consider two demand realizations $d_1$ and $d_2$ to approximate the total reward for different actions, $\overset{*}R(\textbf{I}, \textbf{x},\textbf{d}_i)$ denotes the immediate reward (cost) based on inventory decision $\textbf{x}$ (for example fixed order cost, holding cost, revenue from sales etc.); $V(\mathcal{T}(\textbf{I},\textbf{x},\textbf{d}_i))$ is the value to go estimate (from Eq. \ref{eq:value_to_go_AE}); $\mathcal{L}(\textbf{I})$ denotes the set of feasible inventory actions given the current state \textbf{I}, and $\gamma$ is the pre-selected discounting parameter. Both $\mathcal{L}(\textbf{I})$ and $\textbf{I}^{\prime}$ can be represented using math programming constraints. Furthermore, the set of constraints discussed above can ensure that the value-to-go in the objective function is evaluated correctly. Finally, the optimal action can be computed by solving this IP.
\begin{figure*}[h]
\centering
    \begin{subfigure}[b]{0.38\textwidth}
         \centering
         \includegraphics[width=\textwidth]{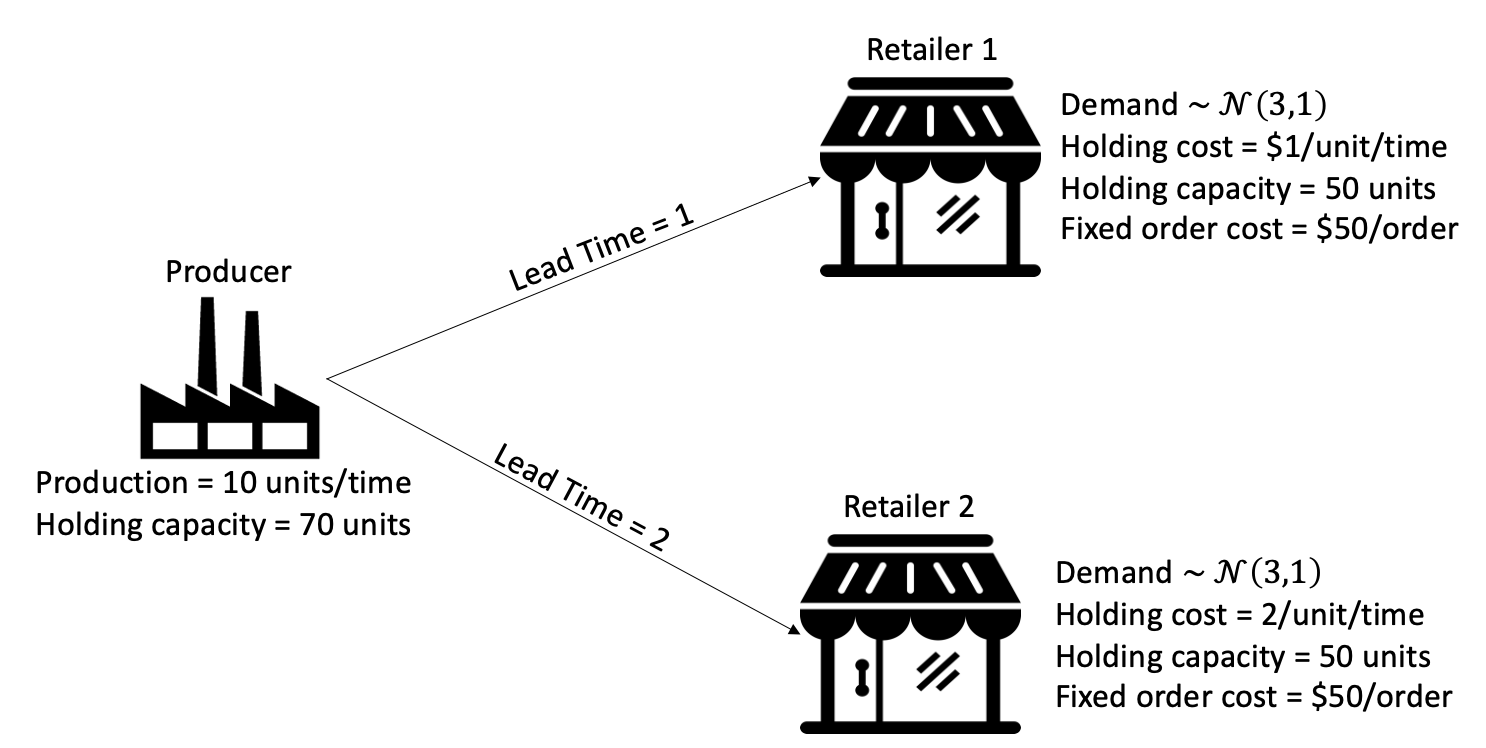}
         \caption{1S-2R}
         \label{fig:NN_small_SC_Example}
     \end{subfigure}
     \begin{subfigure}[b]{0.49\textwidth}
         \centering
         \includegraphics[width=\textwidth]{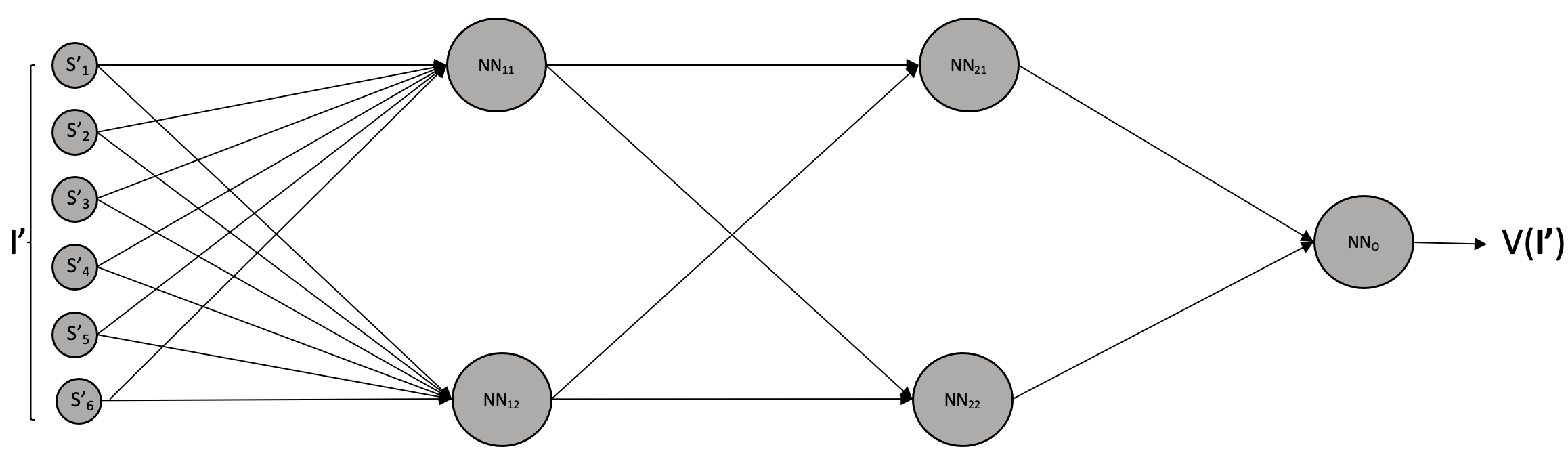}
         \caption{Simple value-to-go neural network}
         \label{fig:NN_small_SC_NN_Example2}
     \end{subfigure}
    \caption{Simplified supply chain network. On the left, we consider a single echelon setting with a single producer and two retailers with lost sales. On the right, a simplified neural network with an input layer (2 neurons), a hidden layer (2 neurons) and an output layer. }
    \label{fig:rate_of_learning}
\end{figure*}
\clearpage
\section{More details on benchmark algorithms and parameter settings}\label{app:benchmark_algos}
\subsection{Supply chain parameters in different settings}\label{app:parameter_table}
\begin{table*}[h]
\centering
\scalebox{0.75}{\parbox{\textwidth}{
\hspace{-1.8cm}
\begin{tabular}{@{}lllllllll@{}}
\toprule[1.0pt]
Parameters & \phantom{a} & 1\texttt{S}-3\texttt{R}-High & 1\texttt{S}-3\texttt{R} & 1\texttt{S}-10\texttt{R} & 1\texttt{S}-20\texttt{R} & 1\texttt{S}-2\texttt{W}-3\texttt{R}  & 1\texttt{S}-2\texttt{W}-3\texttt{R} (DS) & 1\texttt{S}$^{\inf}$-2\texttt{W}-3\texttt{R}  \\ \midrule[1.0pt]
Retailer demand distribution && [N(2,10)] & [N(2,10)] & [N(2,10)] & [N(2,10)]& [N(2,10)]& [N(2,10)]& [N(2,10)]\\\midrule[0.4pt]
Retailer revenue per item            &&    [50]   & [50]     & [50] & [50] & [50] & [50] & [50] \\ \midrule[0.4pt]
Retailer holding cost          & &    [1,2,4]      & [1,2,4]  & [1,2,4,8] & [1,2,4,8]  & [1,2,4]  & [1,2,4] & [1,2,4] \\ \midrule[0.4pt]
Retailer holding capacity && [50] & [50] & [50] & [50] & [50]& [50]& [50]\\\midrule[0.4pt]
Supplier production qty per step && 15 & 10 & 25 & 40 & 10 & 10 & 100 \\\midrule[0.4pt]
Supplier holding capacity && 100 & 100 & 150 & 300 & 100 & 100 & 500 \\\midrule[0.4pt]
Warehouse holding cost     && - & - &-  &-   & [0.5] & [0.5, 0.1]  & [0.5] \\ \midrule[0.4pt]
Warehouse holding capacity && - & - & - & - & [150] & [150] & [150]\\\midrule[0.4pt]
Spillage cost at \texttt{S},\texttt{W},\texttt{R}           &&    [10]      & [10] & [10]  & [10] & [10] & [10] & [10] \\ \midrule[0.4pt]
Lead time (S or W to R)       &&       [1,2,3]  & [1,2,3] & [1,2,3]  & [1,2,3]  &  [1,2,3]& [(1,5),(2,6),(3,7)]  &  [1,2,3]\\ \midrule[0.4pt]
Lead time (\texttt{S} to \texttt{W})       &&   -  & -  & - & - &  [2]&  [2] &  [2] \\ \midrule[0.4pt]
Fixed order cost (\texttt{S} or \texttt{W} to \texttt{R})         &&    [50]      & [50] & [50]  & [50]  & [50] & [50] & [50] \\ \midrule[0.4pt]
Fixed order cost (\texttt{S} to \texttt{W})         &&    -     & - & - & - & [0] & [0] & [0] \\ \midrule[0.4pt]
Variable order cost (any link)         &&    [0]      & [0] & [0] & [0] & [0] & [0] &[20] for \texttt{S}-\texttt{W} \\\midrule[0.4pt]
Maximum order (any link)             &&    [50] & [50]  & [50]    & [50]  & [50] & [50] & [50] \\ \midrule[0.4pt]
\shortstack{Initial inventory distribution\\ (node or link)} && [U(0,4)] & [U(0,4)]  & [U(0,4)] & [U(0,4)]  & [U(0,4)] & [U(0,4)]& [U(0,4)]\\
\bottomrule[1.0pt]
\end{tabular} }}

\caption{Environment parameters for different supply chains studied. In all the settings, we assume deterministic, constant per-period production and that the variability is only in the demand.
The parameters are provided by node type - Retailer (\texttt{R}), Supplier (\texttt{S} or \texttt{S}$^{\inf}$) and Warehouse (\texttt{W}) - and then by links between them. Whenever they are provided in a list format, they correspond to the retailers and warehouses in a chronological order (i.e., \texttt{R}1, \texttt{R2}, \texttt{R3} or \texttt{W1}, \texttt{W2}).
Also, when there are more nodes or links than the parameters (few elements in the list specified in the table), it implies that the parameters list in repeated in a cyclic fashion. For example the lead time (\texttt{S} or \texttt{W} to \texttt{R}) for the environment 1\texttt{S}-10\texttt{R} is given by [1,2,3] and this implies that the lead time for links [\texttt{S}-\texttt{R}1, \texttt{S}-\texttt{R}2,....,\texttt{S}-\texttt{R}10] is (1,2,3,1,2,3,1,2,3,1).
The notation for the distributions used are $N(\mu, \sigma)$ for a normal distribution with mean $\mu$ and standard deviation $\sigma$ and $U(a,b)$ discrete uniform between $a$ and $b$. Note that because demand is discrete and positive, when we use a normal distribution, we round and take the positive parts of the realizations.
The lead-time list has a tuple representation in the dual-sourcing setting to represent the lead time of a retailer from the two different warehouses. For example (1,5) in the list represents the lead time for \texttt{W}1-\texttt{R}1 and \texttt{W}2-\texttt{R}2. Next, we discuss different benchmark algorithms that we tested in this paper.
}\label{table:SCsettings}
\end{table*}
\subsection{Base Stock heuristic policy}~\label{sec:basestock_heuristic}
The seminal work of \citet{scarf1960optimality} shows the optimality of the parametric $(s,S)$ base stock policies for retail nodes with infinite capacity upstream supplier in backordered demand settings. This optimality holds even in the case with fixed costs and constant lead times. In this policy, if ${\bf I}$ is the inventory pipeline vector for a firm, the inventory position is defined as $IP = \sum_{i = 0}^L I^j$, where $L$ is the lead time from the supplier, and the order quantity is $\max\{0, S-IP\}$ as long as $IP<=s$ and 0 otherwise. 

Due to the popularity of these policies, we implement a heuristic base stock policy for the multi-echelon networks we study as follows. At the high level, we construct a base stock policy for every link assuming the upstream entity's supply is unconstrained. For the 2 echelon $1\texttt{S}-n\texttt{R}$ environments, we identify the best base stock policy via grid search for each link using a $1\texttt{S}-1\texttt{R}$ environment with unconstrained supply. The reason we do a grid search is because we are in the lost sales setting. Note that we observe that the constrained supply setting also yields very similar results and hence restrict ourselves to the unconstrained supply setting.
For the 3-echelon environments, we use the same strategy for the $\texttt{W}-\texttt{R}$ links but computing inventory positions $IP$ based on the lead time for that link (note that inventory pipelines can be longer than lead time in the dual sourcing setting). For the $\texttt{S}-\texttt{W}$ links we use environments that treat the warehouse as a retailer with demand equal to the sum of the downstream (lead time) retail demands to find the optimal parameters for that link.

\subsection{Decomposition-Aggregation (DA) heuristic \citep{rong2017heuristics} }\label{sec:DA_heuristic}
In this section, we describe our implementation of the DA heuristic. Note that we adaptation of the DA heuristic to the case of a Normal demand distribution as the authors discuss the method in the case of a Poisson demand distribution.

For 1\texttt{S}-n\texttt{R} environments, for every retailer $r$ compute its respective order up to level $S_r = F^{-1}_{D_r}[\frac{b_r}{b_r+h_r}]$ where $F^{-1}_{D_r}$ is the inverse cumulative demand distribution (cdf) of the random variable $D_r \sim (\mu_r (L_r+1), \sigma_r\sqrt{L_r+1} )$. Here $b_r$ is the retailer revenue per item less the variable ordering cost from the supplier and $h_r$ is the holding cost at the retailer. In every period, the retailer orders $S_r - IP_r$ where $IP_r$ is the retailer's inventory position which is the sum of the retailer's on-hand inventory and that in the pipeline vector.

For the 1\texttt{S}-2\texttt{W}-n\texttt{R} environments, we first decompose by sample paths $1\texttt{S}-1\texttt{W}-1\texttt{R}$.

For each such sample path, we compute $S_r = F^{-1}_{D_r}[\frac{b_r + h_{w_r}}{b_r+h_r}]$ where $F^{-1}_{D_r}$ is the inverse cdf of the random variable $D_r \sim N(\mu_r (L_r+1), \sigma_r\sqrt{L_r+1} )$. Here $b_r$ is the retailer revenue per item less the variable ordering cost from the supplier, $h_r$ is the holding cost at the retailer and $h_{w_r}$ is the holding cost of the warehouse in the sample path of interest.

We then compute $q_{w_r} = F\left[0.5 F^{-1}\left[\frac{b_r}{b_r+h_r}\right]+ 0.5 F^{-1}\left[\frac{b_r}{b_r+ h_{w_r}}\right]\right]$ where $F$ and $F^{-1}$ refers to the cdf and inverse cdf of the standard normal distribution N(0,1). We use this to compute echelon order up to level of the warehouse $S_{w_r} = F^{-1}_{D_{w_r}}[q_{w_r}]$ where $D_{w_r}$ is distributed as N$(\mu_r (L_r+L_w+1), \sigma_r\sqrt{L_r+L_w+1} )$.
An expected shortfall is computed which is $Q_{D_{w_r r}}(s_{w_r}) = E_{D_{w_rr}}[D_{w_rr} - s_{w_r}]^+ $ where $s_{w_r} = S_{w_r}-S_r$.

Next we aggregate across sample paths to recompute the order up to level at common warehouse $w$ using a back-order matching method described as follows: $S_w = Q^{-1}_{D_w}\left(\sum_{r|w_r = w} Q_{D_{w_r r}}(s_{w_r})  \right) $
where $D_w \sim N\left(\sum_{r|w_r = w}\mu_r L_w, \sqrt{\sum_{r|w_r = w}\sigma_r^2 L_w} \right)$ and $Q_{D_{w}}^{-1}(y) = \min \left\{ S | E_{D_{w}}[D_{w} - S]^+ \leq y\right\}$.

In every period, the retailer orders $S_r - IP_r$ from the warehouse and the warehouse orders $S_w - IP_w$ from the supplier where $IP_r$, $IP_w$ are the retailer's and warehouse's respective inventory positions which is the sum of the on-hand inventory and that in the pipeline vector.

\subsection{Parameters and policy for the 1\texttt{S}$^{\inf}$-1\texttt{R} environment}~\label{sec:params1s1r}
The 1\texttt{S}$^{\inf}$-1\texttt{R} environment is a setting with back ordered demand and no-fixed costs. The lead time ($L$) is 4, the holding cost ($h$) is 1.8 and the back order penalty ($b$) is 7. The demand ($D$) is assumed to be normal $N[\mu, \sigma]$ where $\mu = 5$ and $\sigma = 0.8$. In this setting the optimal policy is an order-up to policy with closed form solution $F_{D}^{-1}\left(\frac{b}{b+h}\right)$ which is 26.48. As the environment only allows ordering integer quantities, we tested both 26 and 27 order up-to levels and 27 outperformed on average across 10 (identical) model runs each with 20 test runs with trajectory length of 10K steps.

\subsection{Benchmarking PARL and PPO on serial acyclic chains}\label{app:acyclic}
In this section, we discuss results from additional numerical experiments to test the performance of the proposed PARL algorithm on serial and general-acyclic-non-tree networks. We implement the optimal Clark and Scarf solution\footnote{Using exhaustive search over all possible order-upto integral solutions} for these settings and compare PARL and PPO’s performance against the optimal solution. We test the $1\texttt{S}^{\inf}-1\texttt{W}-1\texttt{W}-1\texttt{R}$, and $1\texttt{S}^{\inf}-1\texttt{W}-1\texttt{W}-1\texttt{W}-1\texttt{R}$ settings with a lead time of 1 between all nodes. The backorder cost is set to 20. The holding costs at the warehouses and retailers are set to 1, 2, and 4, and 1, 2, 3, and 5 respectively. Finally, following standard practice, we also apply holding costs for the in-transit inventory. In Figure \ref{fig:clarkscarf1}, we present results from these experiments. We find similar insights as before: both PARL and PPO’s performance is comparable to the optimal policy, with an optimality gap of 0-5\% in these settings.

\begin{figure*}[h]
\centering

         \centering
         \includegraphics[width=0.6\textwidth]{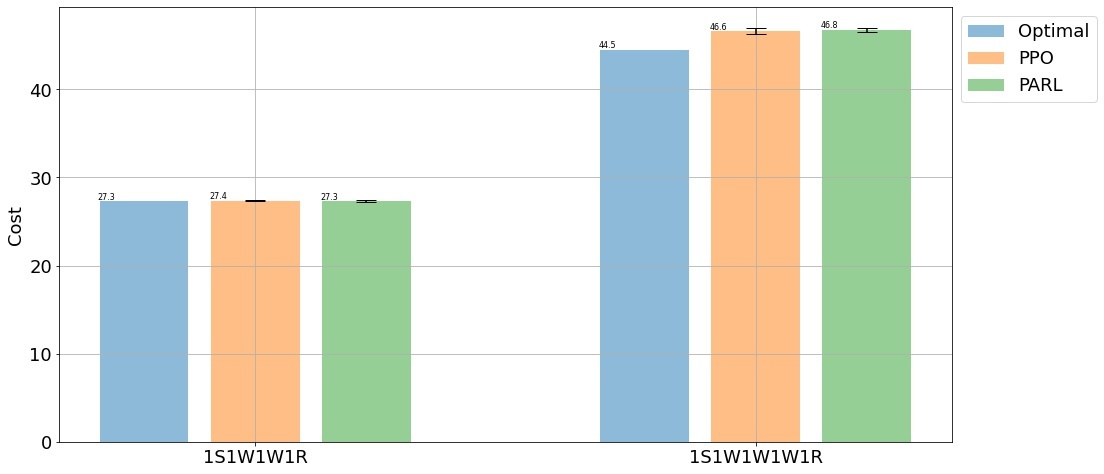}
    \caption{Benchmarking against Clark-and-Scarf policies for serial acyclic supply chains. We consider two different acyclic supply chains ($1\texttt{S}^{\inf}-1\texttt{W}-1\texttt{W}-1\texttt{R}$, and $1\texttt{S}^{\inf}-1\texttt{W}-1\texttt{W}-1\texttt{W}-1\texttt{R}$) and benchmark PARL and PPO against the Clark-and-Scarf based optimal policy. Both PARL and PPO's performance is comparable to the optimal policy's performance in this setting.}
    \label{fig:clarkscarf1}
\end{figure*}

\subsection{Cost and reward comparison for other settings}\label{subsec:other_reward_figs}
\begin{figure*}[h]
\centering
    \begin{subfigure}[b]{0.32\textwidth}
         \centering
         \includegraphics[width=\textwidth]{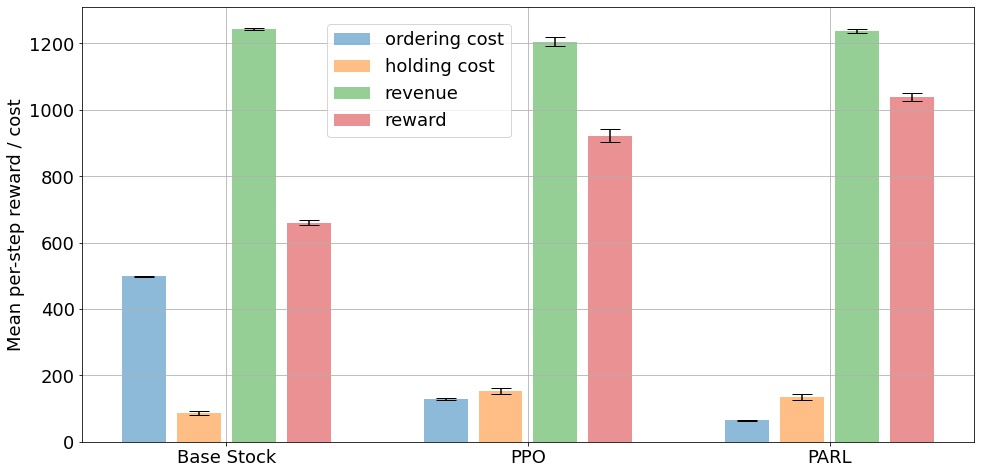}
         \caption{1\texttt{S}-10\texttt{R}}
         \label{fig:cost_1s10r}
     \end{subfigure}
     \begin{subfigure}[b]{0.32\textwidth}
         \centering
         \includegraphics[width=\textwidth]{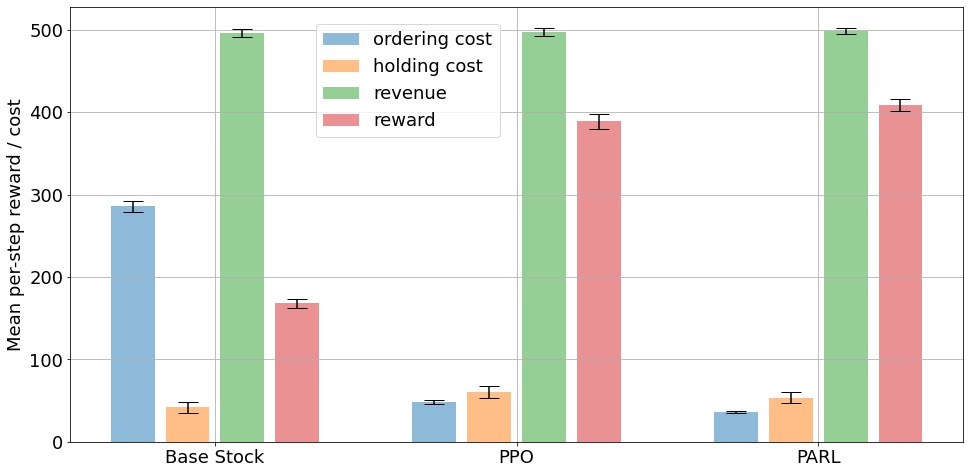}
         \caption{1\texttt{S}-2\texttt{W}-3\texttt{R} (DS)}
         \label{fig:cost_1s2W3r_DS}
     \end{subfigure}
          \begin{subfigure}[b]{0.32\textwidth}
         \centering
         \includegraphics[width=\textwidth]{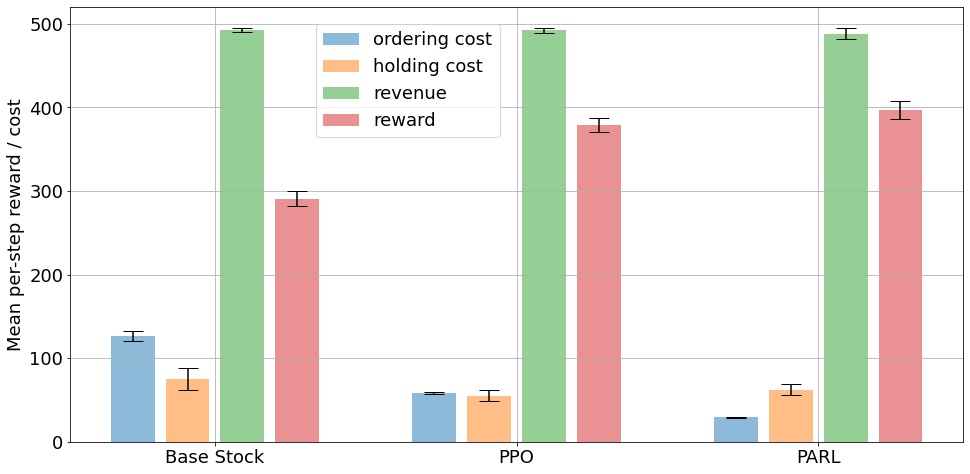}
         \caption{1\texttt{S}-2\texttt{W}-3\texttt{R}}
         \label{fig:cost_1s2W-3R}
     \end{subfigure}
    \caption{Breakdown of rewards in test across BS, PPO and PARL algorithms for settings not discussed in the main text. Note that ordering costs include fixed and variable costs of ordering, revenue refers to the revenue earned from sales and reward refers to the revenue net costs incurred (see \S \ref{sec:inventory} and Table \ref{table:SCsettings} for more details).   }
    \label{fig:barChartOfCosts}
\end{figure*}

\subsection{Comparison of quantile and random sampling in PARL}\label{app:quantile-random}

In many settings, the decision maker might have extra information on the underlying uncertainty $D$. In these cases, one can use specialized weighting schemes to estimate the expected value to go for different actions, given a state.  For example, consider the case when the uncertainty distribution $P(D=d)$ is known and independent across different dimensions. Let $q_1,q_2,..q_{\eta}$ denote $\eta$ quantiles (for example, evenly split between 0 to 1). Also let $F_j ~\&~ f_j, \forall j=1,2..,\texttt{dim}$, denote the cumulative distribution function and the probability density function of the uncertainty $D$ in each dimension respectively. 
Let $d_{ij} = F^{-1}_j(q_i)~ \& ~ \texttt{w}_{ij} = f_j(q_i), \forall i=1,2,..,\eta, j=1,2..,\texttt{dim}$ denote the uncertainty samples and their corresponding probability weights. Then, a single realization of the uncertainty is a \texttt{dim} dimensional vector $d_i = [d_{i1},..,d_{i,\texttt{dim}}]$ with associated probability weight $\texttt{w}^{pool}_i = \texttt{w}_{i1}*\texttt{w}_{i2}..*\texttt{w}_{i,\texttt{dim}}$. With $\eta$ realizations of uncertainty in each dimension, in total there are $\eta^{\texttt{dim}}$ such samples. Let $\mathcal{Q}=\{d_i,\texttt{w}^{pool}_i\}$ be the set of demand realizations sub sampled from this set along with the weights (based on maximum weight or other rules) such that $|\mathcal{Q}| = \eta$. Also let $\texttt{w}_{\mathcal{Q}} = \sum_{i \in \mathcal{Q}}\texttt{w}^{pool}_i$. Then Problem (\ref{eq:SAA}) becomes
\begin{equation}\label{eq:SAA_discrete}
\hat{\pi}^{\eta}_{j}(s) = \arg \max_{a \in \mathcal{A}(s)}\sum_{d_i \in \mathcal{Q}}\texttt{w}_i\left(R(s,a,d_i)+ \gamma \hat{V}^{\pi_{j-1}^{\eta}}_{\theta}(\mathcal{T}(s,a,d_i))\right)\,,
\end{equation}
where $\texttt{w}_i = w^{pool}_i/\texttt{w}_{\mathcal{Q}}$. The computational complexity of solving the above problem remains the same as before but since we use weighted samples, the approximation to the underlying expectation improves. Here we compare the use of quantile sampling and random sampling to generate realizations of the uncertainty in (\ref{eq:SAA}). For the five settings under consideration, we compare the per-step reward and per-step training time across the two sampling approaches. As can be seen in Table \ref{table:quantile-random},  random sampling yields per-step rewards which are close to those obtained via quantile sampling. In terms of training time, random sampling is slower in certain settings (e.g. 1\texttt{S}-3\texttt{R}-High and 1\texttt{S}-10\texttt{R}), with higher per-step train-time average and variance.

\begin{table*}[h]
\centering
\scalebox{0.8}{\parbox{\textwidth}{
\begin{tabular}{@{}lcccccc@{}}
\toprule[1.0pt]
Setting & \shortstack{ PARL-quantile \\ per-step reward} & \shortstack{PARL-random \\ per-step reward} & \shortstack{PARL-quantile \\ per-step train-time (s)} & \shortstack{PARL-random \\ per-step train-time (s)} &  & \\ \midrule[0.5pt]
1\texttt{S}-3\texttt{R}-High	& \shortstack{	514.8	$\pm$	5.3	\\	514.3	}   	& \shortstack{	505.3	$\pm$	11.0	\\	505.1	}   	&	0.178	$\pm$	0.06		   	&  0.457	$\pm$	0.26 	   	&  	&    	\\ \midrule[0.5pt]

1\texttt{S}-3\texttt{R}	& \shortstack{	400.3	$\pm$	3.3	\\	400.8	} 	& \shortstack{	399.5	$\pm$	2.8	\\	400.8	} 	& 	0.051	$\pm$	0.01		 	& 	0.053	$\pm$	0.01		 	& 	&  	\\ \midrule[0.5pt]

1\texttt{S}-10\texttt{R}	& \shortstack{	1006.3	$\pm$	29.5	\\	1015.7	}	& \shortstack{	1005.4	$\pm$	21.1	\\	1007.3	}	&	0.089	$\pm$	0.03			& 	0.12	$\pm$	0.07		& 	& 	\\ \midrule[0.5pt]

1\texttt{S}-2\texttt{W}-3\texttt{R}	& \shortstack{	398.3	$\pm$	2.5	\\	399.7	}	& \shortstack{	395.3	$\pm$	3.1	\\	395.9	}	& 	0.051	$\pm$	0.01		& 	0.050	$\pm$	0.01			&	&  \\ \midrule[0.5pt]

1\texttt{S}-2\texttt{W}-3\texttt{R} (DS)	& \shortstack{	405.4	$\pm$	2.0	\\	405.9	}	& \shortstack{	398.9	$\pm$	9.7	\\	402.0	}	& 	0.044	$\pm$	0.01			& 	0.043	$\pm$	0.01	&	& 	\\ 

\bottomrule[1.0pt]
\end{tabular} }}
\caption{Comparison of quantile and random sampling in PARL.}\label{table:quantile-random}
\end{table*}

\clearpage
\section{Parameter Tuning Details}\label{appendix:parameter_tuning}

\begin{table}[htb]
\caption{Best hyper parameters selected for each environment and method, for the benchmark RL methods.  See Tables \ref{tab:tune_hps_ppo_a2c} and \ref{tab:tune_hps_sac_td3} for hyper parameter abbreviations.}
\label{tab:best_hps}
\begin{center}
\scalebox{0.9}{\parbox{\textwidth}{
\begin{tabular}{|p{0.13\textwidth}|p{0.15\textwidth}|p{0.16\textwidth}|p{0.16\textwidth}|p{0.16\textwidth}| }
\toprule
method &                      SAC &                       TD3 &                      PPO &                     A2C \\
setting           &                          &                           &                          &                         \\
\midrule
1\texttt{S}-3\texttt{R}-High    &    G=0.9 LR=0.01 EO=True &  G=0.9 LR=0.0003 EO=False &   G=0.9 LR=0.003 VFC=1.0 &  G=0.8 LR=0.003 VFC=0.5 \\
\hline
1\texttt{S}-3\texttt{R}         &  G=0.75 LR=0.003 EO=True &  G=0.8 LR=0.0003 EO=False &   G=0.8 LR=0.003 VFC=1.0 &  G=0.8 LR=0.003 VFC=0.5 \\
\hline
1\texttt{S}-10\texttt{R}        &   G=0.8 LR=0.003 EO=True &   G=0.9 LR=0.0003 EO=True &   G=0.8 LR=0.003 VFC=1.0 &  G=0.9 LR=0.003 VFC=1.0 \\
\hline
1\texttt{S}-20\texttt{R}        &   G=0.9 LR=0.003 EO=False &   G=0.75 LR=0.0003 EO=False &   G=0.9 LR=0.010 VFC=3.0 &  G=0.9 LR=0.010 VFC=0.5 \\
\hline
1\texttt{S}-2\texttt{W}-3\texttt{R}      &  G=0.8 LR=0.003 EO=False &  G=0.9 LR=0.0003 EO=False &   G=0.8 LR=0.003 VFC=1.0 &  G=0.8 LR=0.003 VFC=3.0 \\
\hline
1\texttt{S}-2\texttt{W}-3\texttt{R} (DS) &  G=0.9 LR=0.0003 EO=True &   G=0.9 LR=0.0003 EO=True &  G=0.75 LR=0.003 VFC=3.0 &  G=0.8 LR=0.003 VFC=0.5 \\
\hline
1\texttt{S}$^{\inf}$-2\texttt{W}-3\texttt{R} &  G=0.99 LR=0.003 EO=False &   G=0.99 LR=0.003 EO=False &  G=0.99 LR=0.010 VFC=0.5 &  G=0.99 LR=0.003 VFC=3.0 \\
\bottomrule
\end{tabular}}}
\end{center}
\end{table}
  \newcolumntype{P}[1]{>{\centering\arraybackslash}p{#1}}
    \newcolumntype{M}[1]{>{\centering\arraybackslash}m{#1}}
    \begin{table*}[htb]
    \caption{Tuning hyper parameters and additional fixed hyper parameters for PPO and A2C - we vary $\gamma$, learning rate, and value function coefficient - resulting in 36 hyper parameter combinations}
    \label{tab:tune_hps_ppo_a2c}
    \begin{center}
    \scalebox{0.9}{\parbox{\textwidth}{
        \begin{tabular}{|m{0.7\textwidth}|M{0.2\textwidth}| }
         \hline
          \bf Hyper Parameters for PPO and A2C & \bf Value(s) \\
          \hline\hline
          Discount Factor - $\gamma$ \textit{(G)} & 0.99, 0.9, 0.80, 0.75 \\
         \hline
          Learning rate \textit{(LR)} & 0.01, 0.003, 0.0003 \\
         \hline
          Value function coefficient (in loss) \textit{(VFC)} & 0.5, 1.0, 3.0 \\
         \hline
          Number of steps to run per update (epoch length) & 2048 \\
         \hline
          Max gradient norm (for clipping) & 0.5 \\
         \hline
          GAE lambda (trade-off bias vs. variance for Generalized Advantage Estimator) & 0.95 (PPO) and 1.0 (A2C) (defaults)  \\
         \hline
          Number of epochs to optimize surrogate loss (internal train iterations per update - PPO only) & 20 \\
         \hline
          KL divergence threshold for policy update early stopping per epoch (PPO only)  & 0.15 (``target\_kl''=0.1) \\
         \hline
          Clip range (PPO only)  & 0.2 \\
         \hline
          RMSprop epsilon (A2C only) & 1e-05\\
         \hline
        \end{tabular}}}
    \end{center}
    \end{table*}
    \begin{table*}[htb]
    \caption{Tuning hyper parameters and additional fixed hyper parameters for SAC and TD3 - we vary gamma, learning rate, and exploration options - resulting in 32 hyper parameter combinations}
    \label{tab:tune_hps_sac_td3}
    \begin{center}
        \begin{tabular}{|m{0.7\textwidth}|M{0.2\textwidth}| }
         \hline
          \bf Hyper Parameters for SAC and TD3 & \bf Value(s) \\
          \hline\hline
          Discount Factor - $\gamma$ \textit{(G)} & 0.99, 0.9, 0.80, 0.75 \\
         \hline
          Learning rate \textit{(LR)} & 0.01, 0.003, 0.0003, 0.00003 \\
         \hline
          Use generalized State Dependent Exploration vs. Action Noise Exploration (SAC) or Action Noise vs. not (TD3) \textit{(EO)} & True, False \\
         \hline
          Tau (soft update coefficient) & 0.005 \\
         \hline
          Replay buffer size & $10^5$ \\
         \hline
          Entropy regularization coefficient (SAC only) & auto \\
         \hline
        \end{tabular}
    \end{center}
    \end{table*}

        \begin{table*}[htb]
    \caption{Fixed set of hyper-parameters used for all methods}
    \label{tab:fixed_hps}
    \begin{center}
        \begin{tabular}{ |p{0.12\textwidth}|p{0.14\textwidth}|p{0.12\textwidth}|p{0.14\textwidth}|p{0.14\textwidth}|p{0.12\textwidth}| }
         \hline
          Batch size & Net arch. (hidden layers per net) & Activation & State representation & Action representation & Epoch length \\ \hline
         64 & 64x64 & ReLU & continuous (normalized) & continuous (normalized) & 2048 \\ \hline
        \end{tabular}
    \end{center}
    \end{table*}
    \newpage
    \section{Additional experiments to test the scalability of the proposed method}\label{app:runtime_exps}

In this section, we further expand on the computational experiments to test the scalability of the proposed framework. We consider two specific dimensions of the problem: (i) number of retailers in the supply chain network and (ii) number of hidden layers in the value approximating NN. In Figure \ref{fig:scaling_experiments2} we plot the results from these experiments\footnote{We ran these timing calculations on a few different but same-configured linux VMs with Ubuntu 22.04 OS.  Each VM has Intel(R) Xeon(R) Gold 6140 CPUs @ 2.30GHz (i.e., Intel processors) and has 42 cpus and 188GB.
When we ran the experiments, for each run we used 8 processes to parallelize running different trajectories (episodes) and we used 2 threads for solving each MIP.
For each timing experiment, for each value on the x-axis (size) we ran the environment and training for 10 epochs (with exploration amount set at the average across a normal 500 epoch run), and averaged the total time over the 10 epochs to get the average time per epoch.  We repeated this 15 times for each setting as well - to account for differences in randomization and timing for running on the virtual machines (for which timing can change due to network load and traffic).  We report the statistics of the 15 runs for each setting.}. As expected, as the size of the supply chain or the NN grows, so does the runtime of the algorithm. In Figure \ref{fig:scaling_retailers2}, we scale the number of retailers from 5 to up to 320 retailers in the supply chain\footnote{Note that the supply chain network is restricted to the 1S-MR setting. Furthermore, the Neural Network configuration is the same as the rest of the paper: it has two fully connected hidden layers of 64 neurons each. } and analyze the per epoch runtime for different supply chain network sizes.  We find that the runtime scales nearly linearly from 4 seconds per epoch (for the smallest supply chain) to up to 2000 seconds for the largest network. The effect of scaling NN on the runtime is more pronounced. In Figure \ref{fig:scaling_NN2} we scale the number of neuron in the hidden layer from 8 to 512 \footnote{In this case, we fix the supply chain network to be the 1S-3R and scale the number of neurons in each hidden layer of the neural network}. The runtime increases nearly exponentially as the number of neurons in the hidden layer increases. This is as expected since scaling the number of neurons, scales the number of connections in the dense neural network quadratically which translates to bigger IP formulations. Finally, given these insights, we note that moderate to large sized supply chain networks (300 retailers) with moderately sized NN approximator (64 neurons per hidden layer, 2 fully connected hidden layers) can be trained in approximately 10 days in our computational setup. Since all this training can happen offline and the learned model can be directly used for inventory replenishment decisions in real-time. To decrease the compute time further, we note two possible directions for further improving the computations speed: (i) Any episode / trajectory / hyper parameter computation can be done completely in parallel, given enough compute resources - as a training batch is made up of some number of episodes, this parallelization avoids the cost of additional episodes in a batch. (ii) Solving the IP could also be sped up by using more CPUs/parallel processes or threads as well. We used 2 threads in our experiments, and more threads can speed up the process further. Nevertheless, both these directions require additional computational resources that might limit wider adaptability.


\begin{figure*}[h]
\centering
    \begin{subfigure}[b]{0.45\textwidth}
         \centering
         \includegraphics[width=\textwidth]{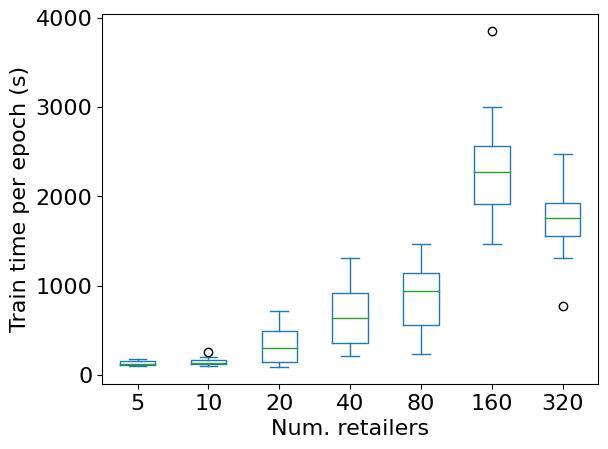}
         \caption{1S-mR}
         \label{fig:scaling_retailers2}
     \end{subfigure}
    \begin{subfigure}[b]{0.45\textwidth}
         \centering
         \includegraphics[width=\textwidth]{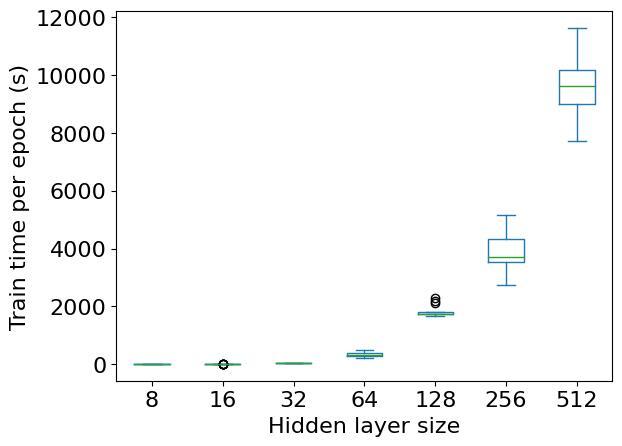}
         \caption{1S-3R}
         \label{fig:scaling_NN2}
     \end{subfigure}
    \caption{Run-time analysis as we scale the supply chain network and the neural network size}
    \label{fig:scaling_experiments2}
\end{figure*}

We also run additional experiments to test how the proposed algorithm scales when using open-source integer programming solvers instead on CPLEX.  In particular, we replicate the IP formulation for neural network value evaluation using both SCIP \citep{BolusaniEtal2024OO} and Pulp-CBC \citep{lougee2003common}. Note that the run-time of the single-step action problem in PARL is driven by the run-time of this evaluation step. Hence, isolating this run-time across different open-source solvers can give us insights on how the per-step action run-time could change if we use open source solvers in our problem. In Figure \ref{fig:runtime_different_solvers}, we compare the solver run-time of both SCIP as well as PuLP-CBC as we increase the size of the neurons in the fully connected hidden layers of the neural network from 8 to 128. As expected, CPLEX performs better than both the open-source solvers. Furthermore, as the complexity of the NN increases, so does the gap between the run-time from CPLEX and the open source solvers. Nevertheless, the difference in runtime between CPLEX and SCIP for moderately sized NNs (up to 64 neurons per layer) is marginal.

\begin{figure*}[h]
\centering
         \includegraphics[width=0.52\textwidth]{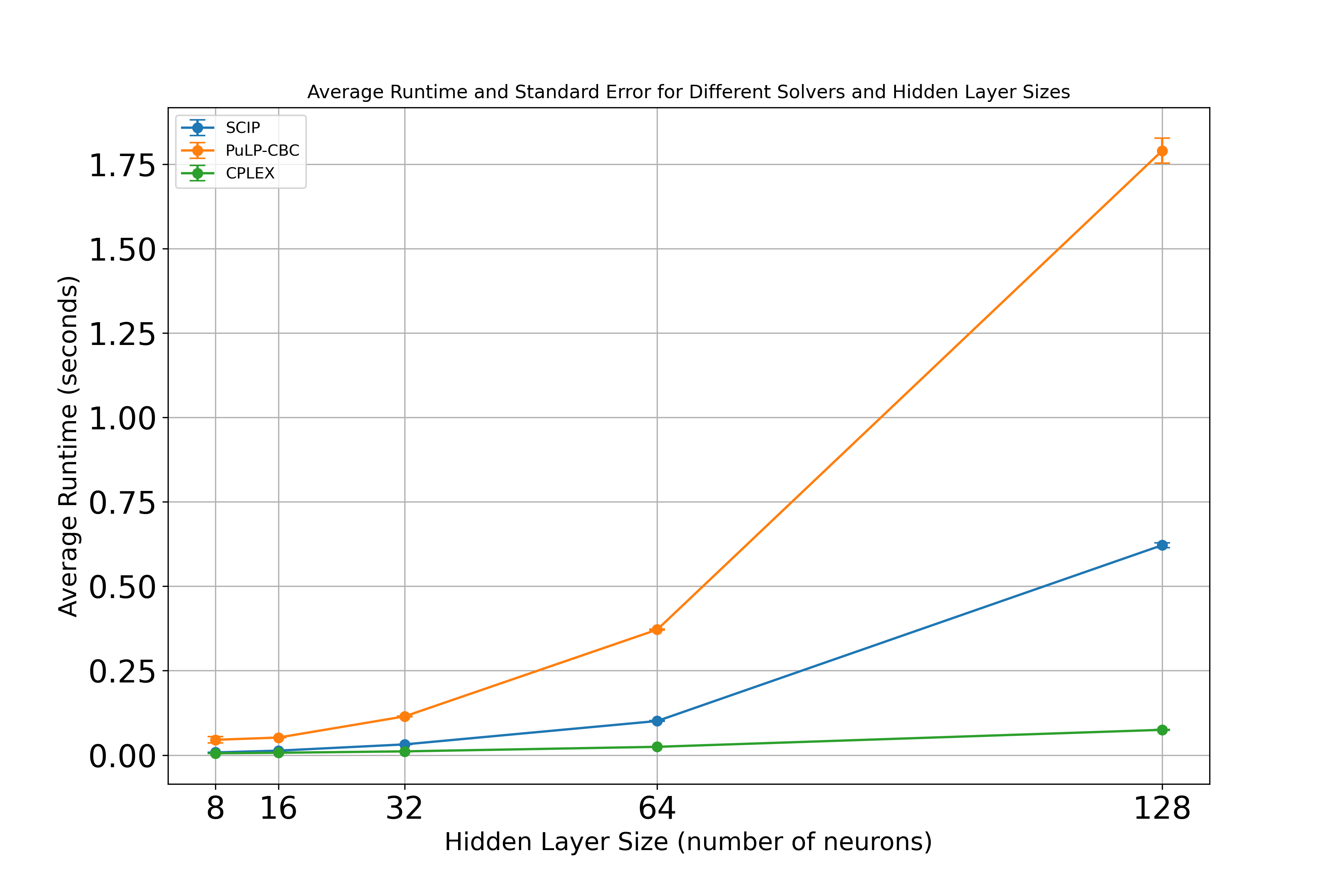}
         \label{fig:NN_small_SC_NN_Example}
    \caption{Run-time analysis as we change the underlying IP solver. }
    \label{fig:runtime_different_solvers}
\end{figure*}
    \section{Open Source Package Usage}\label{app:opensource_listings}
\begin{lstlisting}[label={listing:pdr_1s2w3r_ds_cfg},language=cfg, caption=1\texttt{S}-2\texttt{W}-3\texttt{R} with dual sourcing environment configuration example]
[conf_type]
conf_type = graph  # Specify network via graph or list

[env_params]
env_type = pdr  # pdr (define producers, distributors, and retailers) or 1sMr 
state_rep = N  # Normalized continuous state representation
action_rep = MD  # Multi-discrete action representation
quant = 1  # Order action quantization amount
reset_max_entity_inv = 4  # Max initial inventory randomly generated on reset
reset_max_connection_inv = 4
back_order = False  #Set to True for back order at retailers setting instead of lost sales


[supply_chain_general_params]
max_order_action = 50 # Maximum order amount

# Next, for each of producers, distributors and retailers, define list of IDs and associated lists of settings for each
# Note: default distribution for all entities is (truncated, rounded) stationary Gaussian - other distributions can be easily specified and plugged in (Poisson, non-stationary versions, and inventory-dependent are already available options)

[supply_chain_producer_params]
id_list = P1
prod_daily_prod_avg_list = 10  # Mean production per producer 
prod_daily_prod_std_list = 0. # Std. dev. of production per producer
holding_cost_list = 0
holding_capacity_list = 100
overorder_penalty_list = 0
max_start_inv = -1 # if below 0, set equal to prod_holding_cap

[supply_chain_distributor_params]
id_list = D1, D2
holding_cost_list = 0.5, 0.1
holding_capacity_list = 150, 150
overorder_penalty_list = 10, 10
max_start_inv = 60, 60

[supply_chain_retailer_params]
id_list = R1, R2, R3
demand_avg_list = 2, 2, 2  # Mean demand per retailer 
demand_std_list = 10, 10, 10  # Std. dev. of demand per retailer
revenue_list = 50, 50, 50
holding_cost_list = 1, 2, 4
overorder_penalty_list = 10, 10, 10
holding_capacity_list = 50, 50, 50
max_start_inv = 12, 12, 12
backorder_penalty_list = 0, 0, 0

[supply_chain_connection_params]
# Define connections and their parameters (costs and lead times) between defined entities 
upstream_id_list = P1, P1, D1, D1, D1, D2, D2, D2  
downstream_id_list = D1, D2, R1, R2, R3, R1, R2, R3
L_list = 2, 2, 1, 2, 3, 5, 6, 7  # Specify lead times for each connection
order_cost_per_item_list = 0, 0, 0, 0, 0, 0, 0, 0
order_cost_fixed_list = 0, 0, 50, 50, 50, 50, 50, 50
max_start_inv = 6, 6, 6, 6, 6, 6, 6, 6
\end{lstlisting}

\begin{lstlisting}[label={listing:ex_dnv_usage},language=Python, caption=Example Python code using the library to load a supply chain environment and train and evaluate PARL or a baseline RL algorithm on the environment.]
from supply_chain_rl.run.env_setup_from_cfg import get_env_fun
from stable_baselines3 import PPO
from supply_chain_rl.model.parl import PARL

#Instantiate the environment with normalized continuous action representations
env_function = get_env_fun("env.cfg", action_rep="n")
env = env_function()

#Instantiate a baseline DRL model or PARL model - randomly initialized
# model = PPO(policy="MlpPolicy", env=env)
model = PARL(env=env)

#Fit the model to the env
model.learn(total_timesteps=1e6)

#Evaluate model for some number of steps and log costs and rewards
#Reset environment, send action from model to env to get reward and next state
state = env.reset()
env.logging_enable()

for step in range(256):
    action, _ = model.predict(state)
    state, reward, done, info = env.step(action)
    print(step, reward)

#Export log of evaluation to a file:
env.logging_write_json("trajectory_log.json")
\end{lstlisting}
\end{APPENDICES}
\end{document}